\documentclass{article}

\PassOptionsToPackage{numbers, compress}{natbib}

\usepackage[final]{neurips_2024}

\usepackage{graphicx}
\usepackage[hyphens]{url}

\usepackage{tikz}

\usepackage[utf8]{inputenc}
\usepackage[T1]{fontenc}    

\usepackage{hyperref}       
\usepackage{url}            
\usepackage{booktabs}       
\usepackage{amsfonts}       
\usepackage{nicefrac}       
\usepackage{xcolor}         
\usepackage{multirow}
\usepackage{multicol}
\usepackage{array}
\usepackage{amsmath, amsthm, amssymb, amsfonts}
\newtheorem{definition}{Definition}
\usepackage{mathtools}
\usepackage{wasysym}
\usepackage{bm}
\usepackage{enumitem}
\usepackage{float}
\usepackage{tcolorbox}
\usepackage{tabularx} 
\usepackage{colortbl}
\usepackage{capt-of}
\usepackage{doi}
\usepackage{setspace}
\usepackage{algorithm}
\usepackage[noend]{algpseudocode}
\algrenewcommand{\algorithmicrequire}{\textbf{Input:}}
\algrenewcommand{\algorithmicensure}{\textbf{Output:}}
\usepackage{xspace}
\usepackage{wrapfig,graphbox}
\usepackage[capitalise,nameinlink]{cleveref}
\usepackage{relsize}
\usepackage{caption}
\usepackage{sidecap}
\sidecaptionvpos{figure}{c}
\usepackage{subcaption}

\newcommand{\bgW}{\bm{\mathcal{W}}\xspace}

\DeclareFontFamily{U}{mathx}{}
\DeclareFontShape{U}{mathx}{m}{n}{<-> mathx10}{}
\DeclareSymbolFont{mathx}{U}{mathx}{m}{n}
\DeclareMathAccent{\widecheck}{0}{mathx}{"71}

\NewDocumentCommand{\intz}{o}{
  \IfValueTF{#1}{%
    \tilde{z}_{#1}%
  }{%
    \tilde{\rvz}%
  }%
}
\NewDocumentCommand{\intw}{o}{
  \IfValueTF{#1}{%
    \widetilde{w}_{#1}%
  }{%
    \widetilde{\rvw}%
  }%
}
\renewcommand{\d}{\mathrm{d}}
\newcommand{\vell}{\bm{\ell}}

\newcommand{\verysmallint}{\mathop{\scalebox{0.6}{$\mathlarger{\mathlarger{\smallint}}$}}}

\newcommand{\intu}[1][]{%
    \ifx\relax#1\relax
        \bigocircle[\verysmallint]
    \else
        \bigocircle[\verysmallint \! \! #1]
    \fi
}

\newcommand{\sumu}{\mathop{\bigocircle[+]}}
\newcommand{\produ}{\mathop{\bigocircle[\times]}}
\newcommand{\inpu}{\mathop{\bigocircle[\gaussianbell]}}

\newcommand{\gaussianbell}{%
	\tikz[x=.5em,y=1ex,baseline=-.1em]{%
		\draw[line width=.5pt] plot[domain=-.66:.66] (\x,{1.2*exp(-14*\x*\x)});
	}%
}

\newcommand{\bigocircle}[1][]{%
    \mathop{\mathchoice
        {\tikz[baseline=-0.6ex] \node[circle, draw, inner sep=0pt, minimum size=1.2em] {$\displaystyle #1$};}
        {\tikz[baseline=-0.5ex] \node[circle, draw, inner sep=0pt, minimum size=1.0em] {$\textstyle #1$};}
        {\tikz[baseline=-0.45ex] \node[circle, draw, inner sep=0pt, minimum size=0.9em] {$\scriptstyle #1$};}
        {\tikz[baseline=-0.4ex] \node[circle, draw, inner sep=0pt, minimum size=0.8em] {$\scriptscriptstyle #1$};}
    }
}

\newcommand{\inscope}{\ensuremath{\mathsf{in}}}

\newcommand{\X}{\mathbf{X}}
\newcommand{\x}{\mathbf{x}}
\newcommand{\Y}{\mathbf{Y}}
\newcommand{\y}{\mathbf{y}}
\newcommand{\Z}{\mathbf{Z}}
\newcommand{\z}{\mathbf{z}}

\newcommand{\node}{\mathsf{N}}

\def\eqref#1{equation~\ref{#1}}

\def\1{\bm{1}}

\def\rvb{{\mathbf{b}}}

\def\rvf{{\mathbf{f}}}

\def\rvh{{\mathbf{h}}}
\def\rvu{{\mathbf{i}}}

\def\rvu{{\mathbf{u}}}
\def\rvv{{\mathbf{v}}}
\def\rvw{{\mathbf{w}}}
\def\rvx{{\mathbf{x}}}
\def\rvy{{\mathbf{y}}}
\def\rvz{{\mathbf{z}}}

\def\rmA{{\mathbf{A}}}

\def\rmI{{\mathbf{I}}}

\def\rmW{{\mathbf{W}}}
\def\rmX{{\mathbf{X}}}

\DeclareMathAlphabet{\mathsfit}{\encodingdefault}{\sfdefault}{m}{sl}
\SetMathAlphabet{\mathsfit}{bold}{\encodingdefault}{\sfdefault}{bx}{n}

\def\gL{{\mathcal{L}}}

\def\gR{{\mathcal{R}}}
\def\gS{{\mathcal{S}}}

\def\gU{{\mathcal{U}}}

\def\sR{{\mathbb{R}}}

\def\sX{{\mathbb{X}}}

\usepackage[frozencache,cachedir=.]{minted}
\newminted{python}{%
}

\usepackage{booktabs}
\usepackage{array,multirow,graphicx}
\usepackage{natbib}
\usepackage{enumitem}
\usepackage{xfrac}

\renewcommand{\paragraph}[1]{{\textbf{#1}}}

\hypersetup{
    colorlinks=true,
    linkcolor=teal,
    citecolor=magenta
}

\newcommand{\rgqg}{\text{QG}\xspace}
\newcommand{\rgqt}{\text{QT}\xspace}
\newcommand{\lcp}{\text{CP}\xspace}
\newcommand{\ltucker}{\text{TK}\xspace}

\newcommand{\mnist}{\textsc{Mnist}\xspace}
\newcommand{\fmnist}{\textsc{FashionMnist}\xspace}

\newcommand{\pcmarker}{\protect\tikz \protect\fill[blue] (0,0) circle (0.1cm);}
\newcommand{\PICsharingmarker}{\protect\tikz \protect\fill[green] (0,0) rectangle (0.2cm,0.2cm);}
\newcommand{\PICnosharingmarker}{\protect\tikz \protect\fill[orange] (0,0) -- (0.2cm,0) -- (0.1cm,0.2cm) -- cycle;}

\NewDocumentCommand{\reg}{o}{%
    \IfNoValueTF{#1}{%
        \sX 
    }{%
        \sX^{\scriptscriptstyle{(#1)}}%
    }%
}

\title{Scaling Continuous Latent Variable Models as Probabilistic Integral Circuits}

\author{%
  Gennaro Gala${}^{1,}$ \thanks{
  Corresponding author: \texttt{g.gala@tue.nl} \\
  $\textcolor{white}{.} \, \, \, \, \, \, \intu$ Shared supervision
  }
  \qquad
  Cassio de Campos${}^1$
  \qquad
  Antonio Vergari${}^{2, \intu}$
  \qquad
  Erik Quaeghebeur${}^{1, \intu}$ \\ \\
  ${}^1$Eindhoven University of Technology, NL \\
  ${}^2$School of Informatics, University of Edinburgh, UK
}

\newcommand{\geg}[1]{\textbf{\textcolor{red}{{\textbf{gg}: #1}}}}
\newcommand{\av}[1]{\textbf{\textcolor{orange}{{\textbf{av}: #1}}}}

\renewcommand{\geg}[1]{}
\renewcommand{\av}[1]{}

\begin{document}

\maketitle

\begin{abstract}
    Probabilistic integral circuits (PICs) have been recently introduced as probabilistic models enjoying the key ingredient behind expressive generative models: continuous latent variables (LVs).
    PICs are symbolic computational graphs defining continuous LV models as hierarchies of functions that are summed and multiplied together, or integrated over some LVs.
    They are tractable if LVs can be analytically integrated out, otherwise they can be approximated by tractable probabilistic circuits (PC) encoding a hierarchical numerical quadrature process, called QPCs.

    So far, only tree-shaped PICs have been explored, and training them via numerical quadrature requires memory-intensive processing at scale.
    In this paper, we address these issues, and present:
    (i) a pipeline for building DAG-shaped PICs out of arbitrary variable decompositions,
    (ii) a procedure for training PICs using tensorized circuit architectures,
    and (iii) neural functional sharing techniques to allow scalable training.
    In extensive experiments, we showcase the effectiveness of functional sharing and the superiority of QPCs over traditional PCs.
\end{abstract}

\section{Introduction}

\textbf{\textit{Continuous}} latent variables (LVs) are arguably the key ingredient behind many successful generative models, from variational autoencoders \citep{kingma2013auto} to generative adversarial networks \citep{goodfellow2014generative}, and more recently diffusion models \citep{yang2023diffusion}.
While these models allow to learn expressive distributions from data, they are limited to \textit{sampling} and require task-specific approximations when it comes to perform \textit{probabilistic reasoning}, as even simple tasks such as computing marginals or conditionals are intractable for them.
On the other hand, performing these tasks can be tractable for (hierarchical) \textbf{\textit{discrete}} LV models \citep{choiprobabilistic, choi2011learning}, but these prove to be more challenging to learn at scale \citep{correia2023continuous,dang2022sparse, liu2023scaling, liu2023understanding}.

This inherent trade-off among tractability, ease of learning, and expressiveness can be analyzed and explored with \textit{probabilistic integral circuits} (PICs) \citep{gala24pic}, a recently introduced class of deep generative models defining hierarchies of continuous LVs using \textit{symbolic} functional circuits.
PICs are tractable when their continuous LVs can be analytically integrated out.
Intractable PICs can however be systematically approximated as (tensorized) \textit{probabilistic circuits} (PCs) \citep{vergari2019tractable, choiprobabilistic}, the representation language of discrete LV models.
An instance of such PCs encodes a hierarchical numerical quadrature process of the PIC to approximate, and as such is called \textit{quadrature PC} (QPC).

Distilling QPCs from PICs has proven to be an effective alternative way to train PCs, but it has only been explored for tree-shaped PICs, as building and scaling to richer LV structures is an open research question that requires new tools \citep{gala24pic}.
In this paper, we fill this gap by redefining the semantics of PICs and extending them to DAG-shaped hierarchies of continuous LVs.
Specifically, we design PICs as a language to represent hierarchical quasi-tensors factorizations \citep{townsend2015continuous}, parameterized by light-weight multi-layer perceptrons.

\paragraph{Contributions.}
\textbf{(1)} We present a systematic pipeline to build DAG-shaped PICs, starting from arbitrary variable decompositions (\cref{sec:rg2pic}).
\textbf{(2)} We show how to learn and approximate PICs via a hierarchical quadrature process which we encode in tensorized QPCs that match certain circuit architectures proposed in different prior works \citep{peharz2020random, peharz2020einets, liu2021tractable, loconte2024relationship} (\cref{sec:pic2qpc}).
\textbf{(3)} We present functional sharing techniques to scale the training of PICs, which lead us to parameterize them with multi-headed multi-layer perceptrons (MLPs) requiring comparable resources as PCs (\cref{sec:functional sharing}).
\textbf{(4)} In extensive experiments (\cref{sec:exp}), we show that (i) functional sharing proves remarkably effective for scaling and that (ii) QPCs outperform PCs commonly trained via EM or SGD, while being distilled from PICs with up to 99\% less \textit{trainable parameters}.

\section{Probabilistic integral circuits}
\label{sec:background}

\paragraph{Notation.}
We denote input variables as $\X$ and latent variables (LVs) as $\Y$ and $\Z$, with $\x, \y$ and $\z$ as their realization respectively.
We denote scalars with lower-case letters (e.g., $w \in\sR$), vectors with boldface lower-case letters (e.g., $\rvw \, {\in} \, \sR^N$), matrices with boldface upper-case letters (excluding ${\X, \Y, \Z}$, e.g., $\rmW \in \sR^{M \times N}$), and tensors with boldface calligraphic letters (e.g., $\bgW \in \sR^{\smash{L \times M \times N}}$).

A \textbf{\textit{probabilistic integral circuit}} (PIC) $c$ is a symbolic computational graph representing a non-negative function $c(\rmX) = \smallint c(\X, \rvz) \, \d\z$, i.e.\ $c(\rvx) \geq 0$, over observed variables $\rmX$ and \emph{continuous} latent variables $\Z$.
Similar to probabilistic circuits (PCs) \citep{vergari2019tractable,choiprobabilistic},
PICs have \textit{input}, \textit{sum} and \textit{product} units.\footnote{We refer the reader to \cref{app:circuit} for an introductory overview of (probabilistic) circuits.}
Different from PCs, however, PICs operate on functions, not scalars, and make use of a new type of unit: the \textit{integral} unit $\intu$, which allows to represent continuous LVs.
%
To satisfy non-negativity, we parameterize PICs with non-negative functions and positive sum weights, specifically:

\begin{itemize}[left=0.05cm]
    \item An input unit $u$ (depicted as {$\textcolor{blue}\inpu$} in our figures) represents a possibly non-normalized distribution $f_u(\X_u, \Z_u) \, {\rightarrow} \, \sR^{\smash{+}}$, where $\X_u \, {\subseteq} \, \X$ and $\Z_u \, {\subseteq} \, \Z$;

    \item
    %
    A sum unit $u$ (\textcolor{blue}{$\sumu$}) outputs a weighted sum of the functions it receives from its input units, i.e.\ $g_u(\X_u, \Z_u) \, {=} \, \Sigma_{i \in \inscope(u)} w_i g_i(\X_i, \Z_i)$, where $\inscope(u)$ is the set of units $u$ takes as input, $w_i \, {>} \, 0$, $\X_u \, {=} \, \cup_{i \in \inscope(u)} \X_i$ and $\Z_u \, {=} \,\cup_{i \in \inscope(u)} \Z_i$.
    Similarly,  a product unit $u$ (\textcolor{blue}{$\produ$}) outputs the product of its incoming functions, i.e.\ $g_u(\X_u, \Z_u) \, {=} \, \Pi_{i \in \inscope(u)} g_i(\X_i, \Z_i)$;

    \item Finally, an integral unit $u$ (\textcolor{blue}{$\intu$}) encodes an ``uncountable weighted sum'' whose weights are compactly represented by a function $f_u(\Z_u,\Y_u) \, {\rightarrow} \, \sR^{\smash{+}}$,
    where $\varnothing \, {\neq} \, \Y_u \, {\subseteq} \, \Z$ are the LVs that are being integrated out by $u$, while $\Z_u$ can potentially be empty, i.e.~$\Z_u \, {=} \, \varnothing$.
    The unit receives a function $g_i(\X_i, \Y_u)$ from its only input unit $i$ and outputs the function $g_u(\X_i, \Z_u) \, {=} \, \int_{\Delta} f_u(\Z_u,\rvy_u) \, g_i(\X_i,\rvy_u) \, \d\y_u$, where $\Delta \, {=} \, \mathsf{supp}(\Y_u)$ is the support of $\Y_u$.
    For instance, an integral unit $u$ with $f_u(\{Z\}, \{Y\}) \, {=} \, 2Z - Y^2$ and $\mathsf{supp}(Y)=[-1, 1]$ receiving function $g_i(\{X\},\{Y\}) \,{=} \, X^2-3X+4Y$ would output $g_u(\{Z\},\{X\})=\sfrac{2}{3} \, X(X-3)(6Z-1)$.

\end{itemize}

\begin{wrapfigure}{r}{0.42\textwidth}
\begin{minipage}{0.42\textwidth}
\includegraphics[scale=0.45]{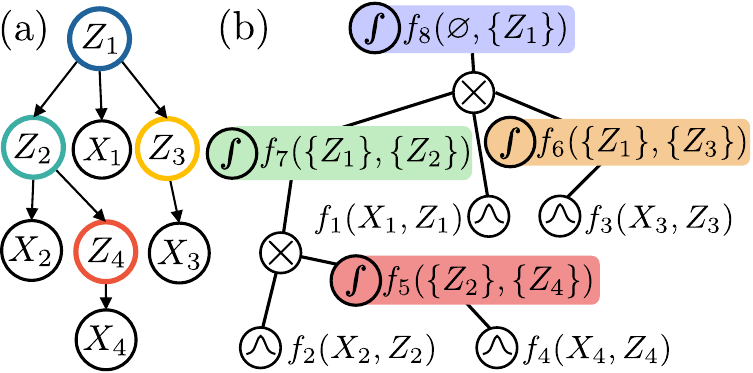}
\caption{PGM (a) $\rightarrow$ tree PIC (b)}
\label{fig:ltm2pic}
\end{minipage}
\vspace{-2.9mm}
\end{wrapfigure}

\cref{fig:ltm2pic}(b) and \cref{fig:rg2pic2qpc}(b) show example PICs.
Note that we use $f$ to indicate the functions attached to input and integral units which are essentially parameters of the model, while we use $g$ to indicate functions being outputted by all type of units.
The output of a PIC $c$ is the output function returned by its root unit $u$, which is only defined on $\X$, i.e.\ $c(\X)$ = $g_u(\X)$ and $\Z_u \, {=} \, \varnothing$.
Similar to PCs, imposing structural constraints over PICs can unlock tractable inference \citep{choiprobabilistic}.
As such, we assume that (i) all $\sumu$-units receive functions defined on the same input variables (aka \textit{smoothness}), and (ii) all $\produ$-units receive functions defined over disjoint sets of input variables (aka \textit{decomposability}).
%

PICs are tractable when their LVs can be analytically integrated out, meaning that we can \textit{pass-through} integral units computing the integration problem they define, eventually outputting a function.
Notably, this is possible when LVs are in linear-Gaussian relationships \citep{koller2009probabilistic} or when functions are polynomials.
Intractable PICs can however be approximated via a hierarchical numerical quadrature process that can be encoded as a PC called \textbf{\textit{quadrature PC}} (QPC).
Intuitively, each PIC integral unit can be approximated by a set of sum units in a QPC, each conditioning on some previously computed quadrature values, with a large but finite number of input units \citep{gala24pic}.
Materializing a QPC allows to train PICs by approximate maximum likelihood:
Given a PIC, gradients to its parameters attached to input, sum and integral units can be backpropagated through the corresponding QPC \citep{gala24pic}.
This also provides an alternative way to train PCs that can rival traditional learners.

So far, the construction of PICs has been limited to a simple \textit{compilation} process from probabilistic graphical models (PGMs) \citep{koller2009probabilistic} with continuous LVs \citep{gala24pic}.
In a nutshell, the LV nodes of a PGM become integral units of a PIC model, and the PGM (conditional) distributions become the functions $f_u$ attached to input and integral units of the PIC, as we illustrate in \cref{fig:ltm2pic}.
However, the PGM structure \textit{needs to be limited to a tree}, as to avoid that the hierarchical quadrature process would yield an exponentially large QPC, thus hindering learning.
This imposes a semantics for current PICs as simple latent tree models \citep{choi2011learning}, and clearly limits their expressiveness as more complex LV interactions are not possible.
Building more expressive PICs requires reinterpreting this semantics and introducing new tools, which we do next.

\section{Building, learning and scaling PICs}

In \Cref{sec:rg2pic}, we systematize the construction of DAG-shaped PICs, showing how to build them starting from arbitrary variable decompositions, going beyond the current state-of-the-art \citep{gala24pic}.
Then, in \Cref{sec:pic2qpc}, we show how to learn and approximate such PICs with QPCs encoding a hierarchical quadrature process, retrieving PC architectures proposed in prior works \citep{loconte2024relationship}.
Finally, in \Cref{sec:functional sharing}, we present (neural) functional sharing, a technique which we use to parameterize PICs as to make their QPC materialization fast and cheap, allowing scaling to larger models and larger datasets.

\subsection{Building PICs from arbitrary variable decompositions}
\label{sec:rg2pic}

Standard PCs can be built according to established pipelines that allow to flexibly represent arbitrary variable decompositions, as well as rich discrete LV interactions \citep{loconte2024relationship, peharz2016latent}.
In the following, we derive an analogous pipeline for PICs that allows to take care of continuous LVs, going beyond their current tree-shaped semantics (\cref{sec:background}) yet allowing to perform hierarchical quadrature without blowing up the size of the materialized QPCs.
To do so, we start by formalizing the notion of hierarchical variable decomposition, or region graph, out of which we will build our PIC structures.

\begin{figure}
    \centering
    \includegraphics[width=\textwidth]{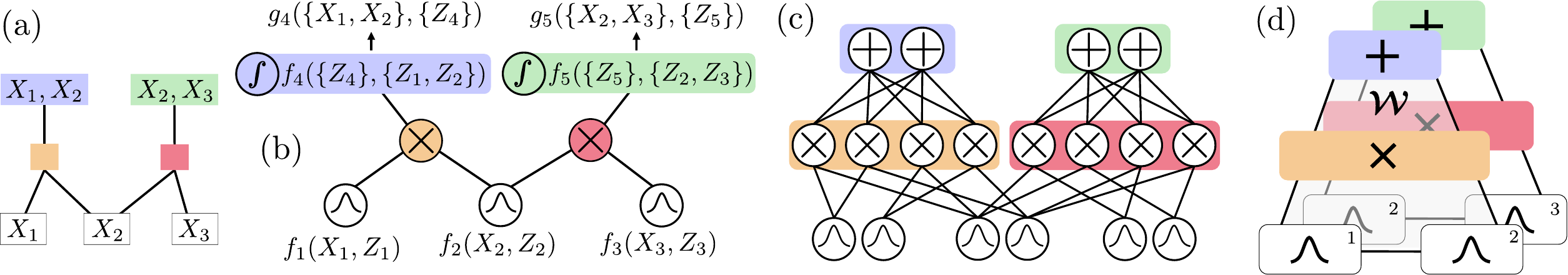}
    \caption{
    \textbf{The pipeline presented in this paper:}
    $\textbf{RG} \rightarrow \textbf{PIC} \rightarrow \textbf{QPC} \rightarrow \textbf{folded QPC}$.
    Starting from a (fragment of) a DAG-shaped region graph (a), we build a DAG-like PIC via \cref{alg:rg2pic} using \textsf{Tucker-merge} (b).
    Then, we materialize a tensorized QPC encoding a hierarchical quadrature process
    via \cref{alg:pic2qpc}, using $K \, {=} \,2$ quadrature points, which we fold to allow faster inference (d).
    }
    \label{fig:rg2pic2qpc}
    \vspace{-4mm}
\end{figure}

\begin{definition}[Region Graph (RG) \citep{dennis2012learning}]
    \label{defn:region-graph}
    An RG $\gR$ over input variables $\X$ is a bipartite and rooted directed acyclic graph (DAG) whose nodes are either \emph{regions}, denoting subsets of $\X$, or \emph{partitions}, specifying how a region is partitioned into other regions (\cref{fig:rg2pic2qpc}(a)).
\end{definition}

RGs can be (i) compiled from PGMs \citep{chow1968approximating, koller2009probabilistic, choi2011learning, liu2021tractable}, (ii) randomly initialized \citep{peharz2020random, di2021random}, (iii) learned from data \citep{dennis2012learning, gens2013learning, molina2018mixed, yang2023bayesian}, or (iv) built according to the data modality (e.g.\ images) \citep{poon2011sum, peharz2020einets, loconte2024relationship}.
%
%
%
If we compile from a tree PGM, as in (\cref{fig:ltm2pic}), the resulting RG will be a tree, thus yielding a tree-like PIC \citep{gala24pic}.
Our pipeline, detailed in \cref{alg:rg2pic}, takes an arbitrary DAG-shaped RG as input, and can deliver DAG-like PICs.
Without loss of generality, we assume to have an RG $\gR$ which only allows for (i) binary partitionings of regions (i.e.\ all product units will have two input units) and (ii) univariate leaves, as shown in \cref{fig:rg2pic2qpc}(a).
Our construction iteratively builds a PIC in a bottom-up fashion, associating regions to PIC units.
For every leaf region $X_u \, {\in} \, \X$ in $\gR$, we instantiate an input unit $u$ with function $f_u(\{X_u\}, \{Z_u\})$, where $Z_u \, {\in} \, \Z$ is an arbitrary continuous LV (\cref{alg-line:pic-input-unit}, \cref{alg:rg2pic}).
Such functions can be univariate conditional densities, i.e.\ $p_u(X_u | Z_u)$, resembling small VAE-like decoders \citep{kingma2013auto} amenable to be numerically integrated.

Once all leaf regions have been processed, we move to the inner ones.
Let $\reg \subseteq \X$ be an inner region partitioned in $N \, {\geq} \, 1$ different ways as $\{(\reg[n]_1, \reg[n]_2)\}_{n=1}^{\smash{N}}$, i.e.\ $(\reg[n]_1 \cap \reg[n]_2) = \varnothing$ and $(\reg[n]_1 \cup \reg[n]_2) = \reg$ for every $n$.
For each partition $(\reg[n]_1, \reg[n]_2)$, we will merge the PIC units associated to regions $\reg[n]_1$ and $\reg[n]_2$ using consecutive applications of product and integral units---as we explain next---eventually associating a unit the partition itself.
One can design such merging as desired, as long as smoothness and decomposability are not violated.
Finally, in case $N \, {=} \, 1$, we associate to $\reg$ the unit associated to its only partition, otherwise, in case $N \, {>} \, 1$, we merge the $N$ units associated to each $n$-th partition using a sum unit which we then associate to $\reg$ (\cref{alg-line:sum-unit}, \cref{alg:rg2pic}).

\paragraph{Merging PIC units.}
Let $u_1$ and $u_2$ be candidate units to merge, each outputting functions with LV $Z_{u_1}$ and $Z_{u_2}$ respectively.
We present two ways of merging units: \textsf{Tucker-merge} and \textsf{CP-merge}, which we detail in \cref{alg:merge} and whose names will be clearer in the next section.
If $Z_{u_1}{\neq}\,Z_{u_2}$, we use \textsf{Tucker-merge}: We merge $u_1$ and $u_2$ with a product, which is then input to an integral unit $u$ with function $f_u(\{Z_u\}, \{Z_{u_1}, Z_{u_2}\})$, where $Z_u \, {\in} \, \Z  \setminus \{Z_{u_1}, Z_{u_2}\}$.
Otherwise, if $Z_{u_1}{=}\,Z_{u_2}$, we use \textsf{CP-merge}: We add two integral units, $u_3$ with input $u_1$ and $u_4$ with input $u_2$, which we finally merge with a product.
We parameterize unit $u_3$ (resp. $u_4$) with $f_{u_3}(\{Z_{u_3}\}, \{Z_{u_1}\})$ (resp. $f_{u_4}(\{Z_{u_4}\}, \{Z_{u_2}\})$), where $Z_{u_3} \, {\neq} \, Z_{u_1}$ (resp. $Z_{u_4} \, {\neq} \, Z_{u_2}$).
Note that whenever merging two units defined on $\X$, we need to marginalize out the remaining LVs, without introducing new ones.
We illustrate the application of \cref{alg:rg2pic} in \cref{fig:rg2pic2qpc}(a-b).
Our pipeline generalizes the PICs used in prior work \citep{gala24pic} (\cref{fig:ltm2pic}) as we can build them by just converting latent tree structures in tree RGs and using $\textsf{CP-merge}$ as merging procedure.
While we now do \textit{not} need a PGM to build a complex PIC structure, one could try to reverse-engineer our PICs to retrieve a PGM via \textit{decompilation} \citep{butz2020sum}, the result would be a very intricate hierarchy over continuous LVs \citep{peharz2016latent}.

\begin{figure}
\begin{minipage}[t]{0.48\linewidth}
\begin{algorithm}[H]
    \small
    \caption{$\textsf{RG2PIC}(\gR)$}\label{alg:overparameterize-tensorize}
    \label{alg:rg2pic}
    $\mathrlap{\textbf{Input}}\phantom{\textbf{Output}}$ RG $\gR$ over variables $\X$ \\
    \textbf{Output} PIC $c(\X) = \int c(\X, \rvz) \d\rvz$
    \setstretch{1.15}
    \begin{algorithmic}[1]
    \State $\gU \leftarrow \textsf{map()}$  \Comment{from regions to PIC units}
    \For{\textbf{each} region $\reg \in \textsf{postOrder}(\gR)$}
        \If{$\reg$ partitioned as $\{(\reg[n]_1, \reg[n]_2)\}_{n=1}^N$}
            \State $\mathrlap{\rho} \phantom{\gS} \leftarrow \mathsf{True}$ \textbf{if} $\reg = \X$ \textbf{else} $\mathsf{False}$
            \State $\gS \! \leftarrow \! \{\textsf{merge}(\gU[\reg[n]_1], \gU[\reg[n]_2], \rho)\}_{n=1}^{N}$
            \State $\gU[\reg] \leftarrow \textsf{pop}(\gS) \, \textbf{if} \; N\mathsf{=}1 \; \textbf{else} \, \sumu([\gS])$ \label{alg-line:sum-unit}
        \Else $\quad \, \triangleright \, |\reg|=1$
            \State $\gU[\reg] \leftarrow \inpu(f_u(\reg, \{Z_u\}))$ \label{alg-line:pic-input-unit}
        \EndIf
    \EndFor
    \State \Return A PIC with $\gU[\X]$ as root unit
    \end{algorithmic}
\end{algorithm}
\end{minipage}
    \hfill
\begin{minipage}[t]{0.48\linewidth}
\begin{algorithm}[H]
    \small
    \caption{$\textsf{merge}(u_1, u_2, \rho)$}
    \label{alg:merge}
    $\mathrlap{\textbf{Input}}\phantom{\textbf{Output}}$ Units $u_1, u_2$, and boolean flag $\rho$ \\
    \textbf{Output} $\intu$ or $\produ$ unit $u$ with $(u_1, u_2)$ as descendants
    \setstretch{1.10}
    \begin{algorithmic}[1]
        \vskip 3.5pt
        \Procedure{}{$\textsf{Tucker-merge}$} \Comment{$Z_{u_1} \neq Z_{u_2}$} 
            \State $\Z_u \leftarrow \varnothing \; \textbf{if} \; \rho \; \textbf{else} \; \{Z\} \subset \Z \setminus \{Z_{u_1}, Z_{u_2}\}$
            \State \Return $\intu(\produ([u_1, u_2]), f_u(\Z_u, \{Z_{u_1}, Z_{u_2}\}))$
        \EndProcedure
        \vskip 2.5pt
        \Procedure{}{$\textsf{CP-merge}$} \Comment{$Z_{u_1} = Z_{u_2}$} 
            \State $\Z_{u_3} \leftarrow \varnothing \; \textbf{if} \; \rho \; \textbf{else} \; \{Z_{u_3}\} \subset \Z \setminus \{Z_{u_1}\}$
            \State $u_3 \leftarrow \intu(u_1, f_{u_3}(\Z_{u_3}, \{Z_{u_1}\}))$
            \State $\Z_{u_4} \leftarrow \varnothing \; \textbf{if} \; \rho \; \textbf{else} \; \{Z_{u_4}\} \subset \Z \setminus \{Z_{u_2}\}$
            \State $u_4 \leftarrow \intu(u_2, f_{u_4}(\Z_{u_4}, \{Z_{u_2}\}))$
            \State \Return $\produ([u_3, u_4])$
        \EndProcedure
    \end{algorithmic}
\end{algorithm}
\end{minipage}
\vspace{-3mm}
\end{figure}

\subsection{Learning PICs via tensorized QPCs}
\label{sec:pic2qpc}

%
%
%
%
Given a DAG-shaped (intractable) PIC,
%
we now show how to approximate it with a tensorized PC encoding a hierarchical quadrature process, namely a QPC.
Intuitively, we interpret PICs as to encode a set of \textit{quasi-tensors} \citep{townsend2015continuous}, a generalization of tensors with potentially infinite entries in each dimension corresponding to a continuous LV, which we materialize into classical tensors via quadrature.
We begin with a definition of tensorized circuits and a brief refresher on numerical quadrature.

\begin{definition}[Tensorized Circuit \citep{loconte2024relationship, peharz2020random}]
    \label{defn:tensorized-circuit}
    A \textbf{tensorized circuit} $c$ is a parameterized computational graph encoding a function $c(\X) \in \sR$, and comprising of \textit{input} $\inpu$, \textit{product} $\produ$ and \textit{sum} $\sumu$ layers.
    Each layer consists of many computational units defined over the same variables, and every non-input layer receives vectors as input from one or more layers.
    Each input layer $\vell$ is defined on variables $\X_{\vell} \, {\subseteq} \, \X$ and computes a collection of $K_{\vell}$ parametric functions $[f_k \, {\colon} \, \mathsf{dom}(\X_{\vell}) \, {\to} \, \sR]_{k=1}^{K_{\vell}}$, outputting a $K_{\vell}$-dimensional vector.
    Each product layer $\vell$ computes either an Hadamard product ($\odot$) or a Kronecker product ($\otimes$) of the vectors it receives from its inputs layers.
    Specifically, the Hadamard product is an element-wise product of vectors, and therefore applicable when these have same size, while the outer product of two vectors $\rvu \, {\in} \, \sR^N$ and $\rvv \, {\in} \, \sR^{M}$ is $\rvw \, {=} \, \rvu \, {\otimes} \, \rvv \, {=} \, ||_{i=1}^N u_i \rvv \in \sR^{\smash{NM}}$, where $||$ is the concatenation operator.
    Finally, a sum layer $\vell$ with $S_{\vell}$ sum units receives inputs from $N$ layers $\{\vell_i\}_{i=1}^N$ and computes the matrix-vector product $\rmW ||_{i=1}^N \vell_i(\X_{\vell_i})$, where $\rmW \in \sR^{S_{\vell} \times K}$, $K{=}\Sigma_{i=1}^{N} K_{\vell_i}$, are the sum layer parameters.
    When $N \, {=} \, 1$, then it simply computes $\rmW \vell_1(\X_{\vell_1})$.
\end{definition}

\paragraph{Numerical quadrature.}
A {numerical quadrature rule} is an approximation of the definite integral of a function as a weighted sum of function evaluations at specified points \citep{davis2007methods}. 
Specifically, given some integrand $f \, {:} \, \sR \, {\rightarrow} \, \sR$ and interval $\Delta \, {:=} \, [a ,b]$, a quadrature rule consists of a set of $K$ integration points $\intz \, {\in} \, \Delta^{\smash{K}}$ and weights $\intw \, {\in} \, \sR^{\smash{K}}$ minimizing the integration error $\varepsilon_K = | \smallint_{\Delta} f(z) \, \mathrm{d}z - \Sigma_{k=1}^{\smash{K}} \intw[k] f(\intz[k]) |$, which goes to zero as $K{\rightarrow} \, \infty$.
To approximate an integral of an $N$-dimensional function $f$, we can phrase the multiple integral as repeated one-dimensional integrals by applying Fubini's theorem \citep{fubini1907sugli}, aka tensor product rule, as follows.
\begin{equation}
    \label{eq:fubini}
    \int_{\Delta^{K}} \! f(\z) \d\z =
    \int_{\Delta} \! ... \Big(\int_{\Delta} \! f(z_1, ..., z_N) \d z_1\Big) ... \d z_N \approx
    \!\! \sum_{i_1 \in [K]} \!\! \intw[i_1] ... \!\! \sum_{i_N \in [K]} \!\! \intw[i_N] f( \intz[i_1], ..., \intz[i_N]).
\end{equation}

\begin{figure}
    \centering
    \includegraphics[width=\textwidth]{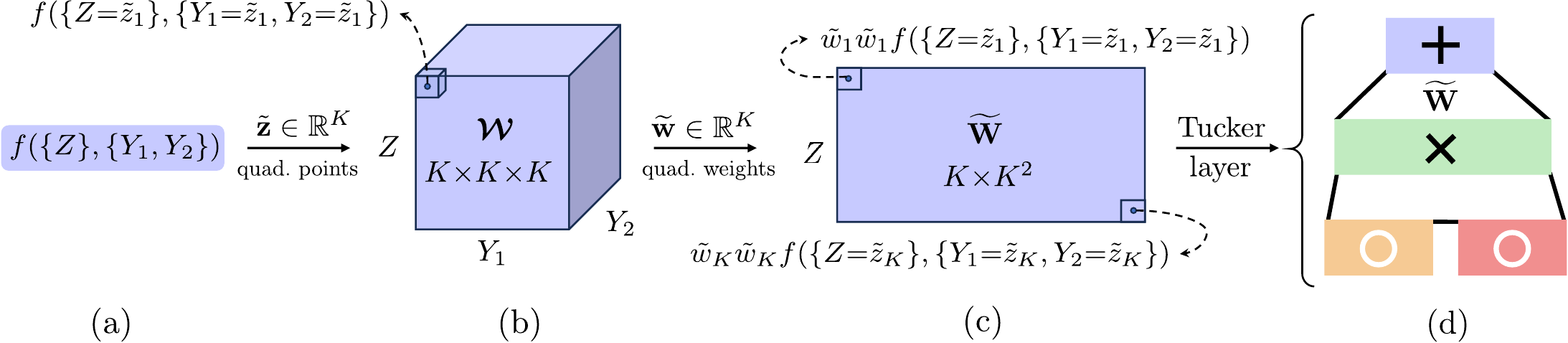}
    \caption{\textbf{From functions to sum-product layers via multivariate numerical quadrature} (\cref{sec:pic2qpc}).
    We illustrate how the 3-variate function $f(\{Z\}, \{Y_1, Y_2\})$ (a) can be seen as an infinite (quasi) tensor that we first materialize w.r.t.\ integration points $\intz$ as a finite tensor $\bgW$ of size $K \times K \times K$ (b, \Cref{eq:tensor-materialization}), then flatten as a matrix accounting for integration weights $\intw$ (c, \Cref{eq:quad-weights}), and finally use to parameterize a Tucker layer (d, \Cref{eq:tucker-layer}).
    }
    \label{fig:quad-tucker}
    \vspace{-3mm}
\end{figure}

\paragraph{From PICs to QPCs.}
Given a candidate PIC, we will explore it in post order traversal, and iteratively associate a circuit layer (\cref{defn:tensorized-circuit}) to each PIC unit, a process that we call \emph{materialization}.
We detail such procedure in \cref{alg:pic2qpc}, which essentially applies \cref{eq:fubini} hierarchically over the PIC units.
Each unit encodes a function over potentially continuous and discrete variables, hence representing a quasi-tensor,
which we approximate by evaluating it over the quadrature points only, thus materializing a classical tensor (\cref{fig:quad-tucker} (a,b)).
We facilitate quadrature by assuming that all PIC LVs have bounded domain $\Delta \, {:=} \, [-1, 1]$.
This way, we can always use the same quadrature rule $(\intz, \intw)$ for each required (multivariate) approximation, and also simplify treatment and exposition.

\begin{wrapfigure}{r}{0.42\textwidth}
\algrenewcommand\algorithmicindent{0.8em} 
\vspace{-7.8mm}
\begin{minipage}{0.42\textwidth}
\begin{algorithm}[H]
    \small
    \caption{$\textsf{PIC2QPC}(c, \intz, \intw)$}
    \label{alg:pic2qpc}
    \textbf{Input} PIC $c$, quadrature rule $\intz, \intw \in \sR^K$ \\
    \textbf{Output} Tensorized QPC
    \begin{algorithmic}[1]
    \setstretch{1.15}
    \State $\gL \leftarrow \textsf{map()}$  \Comment{from PIC units to layers}
    \For{\textbf{each} unit $u \in \textsf{postOrder}(c)$}
        \If{$u$ is $\inpu$}
            \State $\gL[u] \leftarrow [ \, \inpu(f_u(X_u, Z_u{\mathsf{=}} \intz[k])) \, ]_{k=1}^{K}$ \label{alg-line:qpc-input-layer}
        \ElsIf{$u$ is $\intu(u_1, f_u)$} 
            \State $\gL[u] \leftarrow \widetilde{\rmW}^{\scriptscriptstyle (u)} \, \gL[u_1]$ via \cref{eq:quad-weights} \label{alg-line:qpc-sum-layer}
        \ElsIf{$u$ is $\sumu([u_i, w_i]_{i=1}^{N})$}
            \State $\gL[u] \leftarrow \rmW^{\smash{(u)}} ||_{i=1}^{N} \gL[u_i]$ via \cref{eq:mixing-layer} \label{alg-line:qpc-mixing-layer}
        \ElsIf{$u$ is $\produ([u_1, u_2])$}
            \State $\Circle \leftarrow \odot \; \textbf{if} \; \Z_{u_1}{=}\; \Z_{u_2} \; \textbf{else} \; \otimes$
            \State $\gL[u] \leftarrow \gL[u_1] \, \Circle \, \gL[u_2]$
        \EndIf
    \EndFor
    \State \Return A QPC with $\gL[c]$ as root layer
    \end{algorithmic}
\end{algorithm}
\end{minipage}
\vspace{-5mm}
\end{wrapfigure}

We begin materializing every PIC input unit $u$ with function $f_u(\{X_u\}, \{Z_u\})$ w.r.t.\ integration points $\intz$, effectively creating an input layer $\vell \, {:} \, \mathsf{dom}(X_u) \, {\rightarrow} \, \sR^{\smash{K}}$ as $[ \, f_u (X_u, Z_u {=} \intz[k]) \, ]_{k=1}^{\smash{K}}$ (Line 4, \cref{alg:pic2qpc}).
The parameters of such layer can be materialized as a matrix of shape $K \, {\times} \, P$, where $P$ is the number of parameters $f_u$ requires.
For example, if $f_u$ is a univariate conditional Gaussian $p_u(X_u | Z_u)$, we use a $K \, {\times} \, 2$ matrix for parameterizing the layer, where each row stores the mean and standard deviation at each integration point $\intz[k]$.

Next, we address the most important part of this quadrature process, i.e.\ the materialization of PIC integral units as sum layers.
Specifically, let $u$ be an integral unit with $N$-dimensional function $f_u(\Z_u, \Y_u)$, where $|\Z_u| \, {=} \, N_Z, |\Y_u| \, {=} \, N_Y$ and $N \, {=} \, N_{Z} \, {+} \, N_{Y}$.
We materialize $f_u$ w.r.t.\ integration points $\intz$, effectively creating an $N$-dimensional tensor $\bgW{}^{\smash{(u)}} \in \sR{}^{\smash{K \times \dots \times K}}$, such that
\begin{align}
    w_{i_1, \dots, i_N}^{(u)} = f_u (\{Z_{1} {=} \intz[i_1], \dots, Z_{N_{Z}} {=} \intz[i_{N_{Z}}] \}, \{Y_{1} {=} \intz[i_{1+N_{Z}}], \dots, Y_{N_{Y}} {=} \intz[i_N]\}).
    \label{eq:tensor-materialization}
\end{align}
After materializing tensor $\bgW{}^{(u)}$, we flatten it w.r.t.\ variables $\Z_u$ and $\Y_u$, so as creating a matrix $\rmW^{\smash{(u)}}$ of size ${K^{\smash{N_Z}} \times K^{\smash{N_Y}}}$, an operation aka \emph{matricization}.
As last step, we plug-in the quadrature weights $\intw \in \sR^{\smash{K}}$ in $\rmW^{\smash{(u)}}$, arriving to matrix $\widetilde{\rmW}^{\smash{(u)}}$ of ${K^{\smash{N_Z}} \, {\times} \, K^{\smash{N_Y}}}$, whose $i$-th row is
\begin{align}
    \widetilde{\rvw}^{(u)}_{i:} = (\intw \otimes \dots \otimes \intw) \, \rvw^{(u)}_{i:} = \intw^{\otimes N_Y} \, \rvw^{(u)}_{i:},
    \label{eq:quad-weights}
\end{align}
where $\intw^{\otimes N_Y}$ is the vector of size $K^{\smash{N_Y}}$ resulting from the $N_Y$-times application of the Kronecker product $\otimes$ over $\intw$ (Line 6, \cref{alg:pic2qpc}). 
We illustrate this process in \cref{fig:quad-tucker}(a-c).
Similarly, we also materialize every PIC sum unit $u$ with weights $\{w_i\}_{i=1}^{\smash{N}}$ as a sum layer, but parameterized by
\begin{align}
\label{eq:mixing-layer}
\rmW^{\smash{(u)}} = ||_{i=1}^{N} w_i\,\rmI_K \in \sR^{K \times NK},
\end{align}
where $\rmI_K$ is the $K \times K$ identity matrix and $||$ the concatenation operator (Line 8, \cref{alg:pic2qpc}). 
Note that such sum layer can be seen as a \emph{mixing layer} \citep{peharz2020einets, loconte2024relationship}.
Finally, consider a PIC product unit $u$ with inputs $u_1$ and $u_2$, each outputting functions with LVs $Z_{u_1}$ and $Z_{u_2}$ respectively.
We associate to $u$ an Hadamard product layer if $Z_{u_1} \, {=} \, Z_{u_2}$, or a Kronecker product layer if $Z_{u_1} \, {\neq} \, Z_{u_2}$, reflecting the fact that we are marginalizing out two different LVs.
We summarize our PIC materialization---illustrated in \cref{fig:rg2pic2qpc}---in \cref{alg:pic2qpc}, where we iteratively associate a PIC unit to a circuit layer, eventually delivering a tensorized QPC.
We stress that being QPCs just standard PCs they enjoy their same properties (e.g. tractable marginalization).
We will learn PICs via maximizing the likelihood of its QPC materialization.

\paragraph{QPCs as existing tensorized architectures.}
Materializing PICs built via \cref{alg:rg2pic} delivers tensorized PCs with alternating sum and product layers, aka \textbf{\textit{sum-product layers}} \citep{loconte2024relationship}.
An instance of such layers is the Tucker layer, used in architectures like RAT-SPNs \citep{peharz2020random} and EiNets \citep{peharz2020einets}.
Specifically, a binary Tucker layer $\vell$ \citep{tucker64extension} computes
\begin{align}
    \label{eq:tucker-layer}
    \vell(\X_{\vell}) = \rmW \left( \vell_1(\X_{\vell_1}) \otimes \vell_2(\X_{\vell_2}) \right), \tag{Tucker-layer}
\end{align}
where $\rmW\in\sR^{K\times K^2}$ and $\vell_1,\vell_2$ are input layers of $\vell$, each outputting a $K$-dimensional vector.
In contrast, the recent HCLT architectures \citep{liu2021tractable} use the canonical polyadic (CP) layer $\ell$ \citep{carroll1970indscal}, i.e.\
\begin{align}
    \label{eq:candecomp-layer}
    \vell(\X_{\vell})= \left( \rmW^{(1)} \vell_1(\X_{\vell_1}) \right) \odot \left( \rmW^{(2)} \vell_2(\X_{\vell_2}) \right), \tag{CP-layer}
\end{align}
where $\rmW^{(1)}, \rmW^{(2)} \in \sR^{K \times K}$.
We exactly recover Tucker (resp. CP) layers in our QPCs when these are materialized from PICs built via \textsf{Tucker-merge} (resp. \textsf{CP-merge}) in \cref{alg:merge},
and hence the name of the merging procedure.
Therefore, some QPCs can exactly match existent tensorized architectures, and this certainly happens when these are materialized from PICs built via \cref{alg:rg2pic}.
This gives a new point of view on traditional tensorized architectures, and new possibilities for representation learning \citep{vergari2018sum}.
\Cref{fig:quad-tucker} illustrates how the materialization of a 3-variate function leads to a Tucker layer.
This 1-to-1 mapping between tensorized PC architectures and QPCs will allow for a fair comparison in our experiments.

\paragraph{Folding tensorized circuits for faster inference.}
The layers of a tensorized circuit that (i) share the same functional form and that (ii) can be evaluated in parallel, can be stacked together as to create a \emph{folded layer} \citep{loconte2024relationship, peharz2020einets} 
which speeds up inference and learning on GPU by orders of magnitude.
For instance, let $\{\vell_i\}_{i=1}^{F}$ be $F$ parallelizable Tucker layers each parameterized by a matrix $\rmW^{\smash{(i)}}$ of size ${K {\times} K{}^2}$.
Such layers can be evaluated as a folded layer $\vell$ parameterized by a tensor $\bgW$ of size ${F {\times} K {\times} K{}^2}$, which computes the---otherwise sequential---$F$ tucker layers in parallel.
We illustrate folding in \cref{fig:rg2pic2qpc}(d), and later on in \cref{fig:c-sharing}(c).
Note that (i) the input layers sharing the same function form can always be folded and that (ii) although a tensorized circuit may have many types of sum-product layers, using one type only is common in practice, and promotes depth-wise folding.

\subsection{Scaling PICs with neural functional sharing}
\label{sec:functional sharing}

%
%
Materializing QPCs can be memory intensive and time consuming, depending on: (i) the cost of evaluating the functions we need to materialize, (ii) the degree of parallelization of the required function evaluations, and (iii) the number of integration points $K$.
%
To solve these issues,
we introduce \textit{neural functional sharing} \citep{segal2005learning},
i.e.\ we share multi-layer perceptrons as to parameterize multiple PIC units at once.
This allows us to scale to larger models and datasets, as we make materialization faster and more memory-efficient than previous work \citep{gala24pic}.

\begin{figure}
    \centering
    \includegraphics[width=\textwidth]{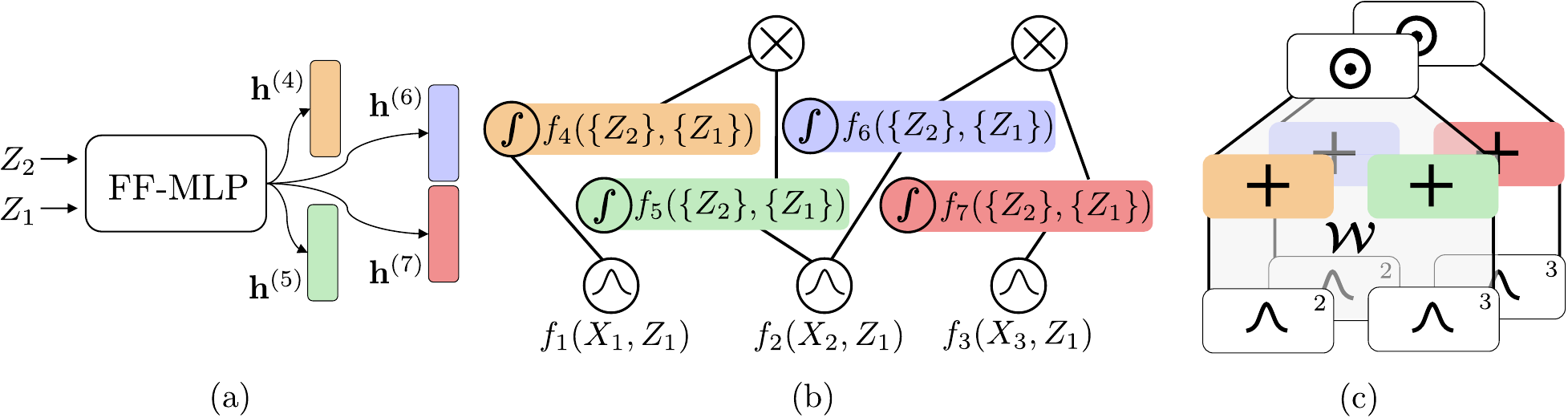}
    \caption{\textbf{From neural C-sharing to folded CP-layer} (\cref{sec:functional sharing}).
    We sketch a 4-headed MLP with Fourier-Features (a) which we use to parameterize a group of 4 integral units (at the same depth) of a PIC (b), whose materialization leads to a folded CP-layer parameterized by a tensor $\bgW$ of size $2 \times 2 \times K \times K$  (c), with $K$ being the number of integration point.
    Note that, during materialization, the FF-MLP block in (a) will be only evaluated $K^2$ times, and not $4K^2$.
    }
    \label{fig:c-sharing}
    \vspace{-3mm}
\end{figure}

\paragraph{PIC functional sharing.}
Functional sharing is to PICs as parameter-sharing is to PCs.
This type of sharing can be applied over a \emph{group} of input/integral units---grouped according to some criteria---whose functions have all the same number of input and output variables.
Specifically, let $\gamma = \{u_i\}_{i=1}^{\smash{N}}$ be a group of $N$ input/integral units, each with function $f_i : \sR^{\smash{I}} \rightarrow \sR^{\smash{O}}$.
The simplest form of functional sharing is to set all functions to be equal, i.e.\ $\forall i,j \, {\in} \, [N]: f_i=f_j$.
In this way, we reduce the number of function evaluations from $NK$ to $K$ as long as we materialize each $f_i$ w.r.t.\ the same integration points $\intz \in \sR^{\smash{K}}$, which is the case for \Cref{alg:pic2qpc}.
We call this type of sharing \textbf{F-sharing}, as per \emph{full-sharing}.
More interestingly, leveraging functional composition, we may define $f_i=h_i \circ f$, so as sharing an \emph{inner} function $f$ for all unit functions.
Similarly as before, as long as we materialize each $f_i$ w.r.t.\ the same quadrature points $\intz \in \sR^{\smash{K}}$, we would only need $K$ function evaluations for $f$ instead of $NK$, as we can share them with all \emph{outer} functions $h_i$ for further evaluation. 
We call this type of sharing \textbf{C-sharing}, as per \emph{composite-sharing}.
The original implementation of PICs \citep{gala24pic} used neither F-sharing nor C-sharing.

Finally, we present and apply two different ways of grouping units.
The first consists of grouping all input units, a technique which is only applicable when all input variables share the same domain.
With this grouping, coupled with F-sharing, we would only need to materialize $K \, {\times} \, P$ parameters, and use them to parameterize every QPC input layer. 
The second consists of grouping all integral units at the same depth of the PIC structure, which we couple with C-sharing and materialize as a folded sum-product layer.
Despite grouping units that materialize into a folded layer is a natural and convenient choice, note that we can also group units that do not materialize as such. 
Once all units in a PIC have been grouped, materialization can be performed per-group.

\paragraph{PIC functional sharing with (multi-headed) MLPs.}
Similar to \citep{gala24pic}, we parameterize PIC input and integral units with light-weight multi-layer perceptrons (MLPs).
However, instead of using a single MLP for each function, we will apply functional sharing as we strive to make the QPC materialization faster and memory efficient.
Specifically, consider a group of integral units $\gamma \, {=} \, \{u_i\}_{i=1}^{\smash{N}}$, each with function $f_i: \sR^{\smash{I}} \rightarrow \sR$, over which we want to apply functional sharing.
For every group $\gamma$, we would have an $L+1$ layered MLP of the form:
\begin{align}
    \phi^{(\gamma)} : \sR^I \rightarrow \sR^M := \phi_{L}^{(\gamma)} \circ \dots \circ \phi_{1}^{(\gamma)} \circ \mathsf{FF},
\label{eq:group-mlp}
\end{align}
where $\mathsf{FF}: \sR^{\smash{I}} \rightarrow \sR^{\smash{M}}$ is a Fourier-feature layer \citep{tancik2020fourier}, and each $\phi_i^{(\gamma)}: \sR^{M} \rightarrow \sR^{M}$ is a standard linear layer followed by an element-wise non linearity $\psi$, i.e.\ $\psi(\rmA \rvz + \rvb)$, with $\rmA \in \sR^{M \times M}, \rvb \in \sR^M$, and $M$ being the size of the MLP.
Applying F-sharing over $\gamma$ would simply consist of setting
\begin{align}
    f_i : \sR^M \rightarrow \sR := \mathsf{softplus}(\rvh^{(\gamma)} \cdot \phi^{(\gamma)} + b^{(\gamma)}),
    \tag{neural F-sharing}
\end{align}
where $\rvh^{(\gamma)} \in \sR^M$ and $b^{(\gamma)} \in \sR$  are group-dependent parameters, therefore making all functions in the group equal.
Instead, to implement C-sharing, we parameterize each $f_i$ as
\begin{align}
    f_i : \sR^M \rightarrow \sR := \mathsf{softplus}(\rvh^{(i)} \cdot \phi^{(\gamma)} + b^{(i)}),
    \tag{neural C-sharing}
\end{align}
where $\rvh^{(i)} \in \sR^M$ and $b^{(i)} \in \sR$ are function-dependent parameters, effectively creating a multi-headed MLP.
%
As an example, consider a folded CP-layer with $F\,{=}\,500$ and $K\,{=}\,512$---which we actually used in practice---resulting in $2FK^2\approx262$M trainable parameters.
Assuming no bias term, an MLP with $L\,{=}\,2$ and $M\,{=}\,256$ would only instead require $LM^2+2FM\approx387$K trainable parameters to materialize the same tensor, resulting in more than 99\% less trainable parameters.
We illustrate such C-sharing in \cref{fig:c-sharing}.
In \cref{app:mlp-details} we provide more details about our MLPs.

\paragraph{Fast \& memory-efficient QPC materialization.}
Combing PIC functional sharing and per-group materialization allows scaling the training of PICs via numerical quadrature, as we drastically reduce the number and the cost of function evaluations required for the QPC materialization.
We can now materialize very large QPCs, matching the scale of recent over-parameterized PCs yet requiring up to 99\% less trainable parameters when using a large $K$.
This was not possible in the original formulation of PIC \citep{gala24pic} as (i) the entire QPC was materialized in one-shot, not per-group, and (ii) no functional sharing was implemented, as each input/integral function had its own MLP.

\section{Experiments}
\label{sec:exp}

\sidecaptionvpos{figure}{c}
\begin{SCfigure}
    \includegraphics[align=c, scale=0.37]{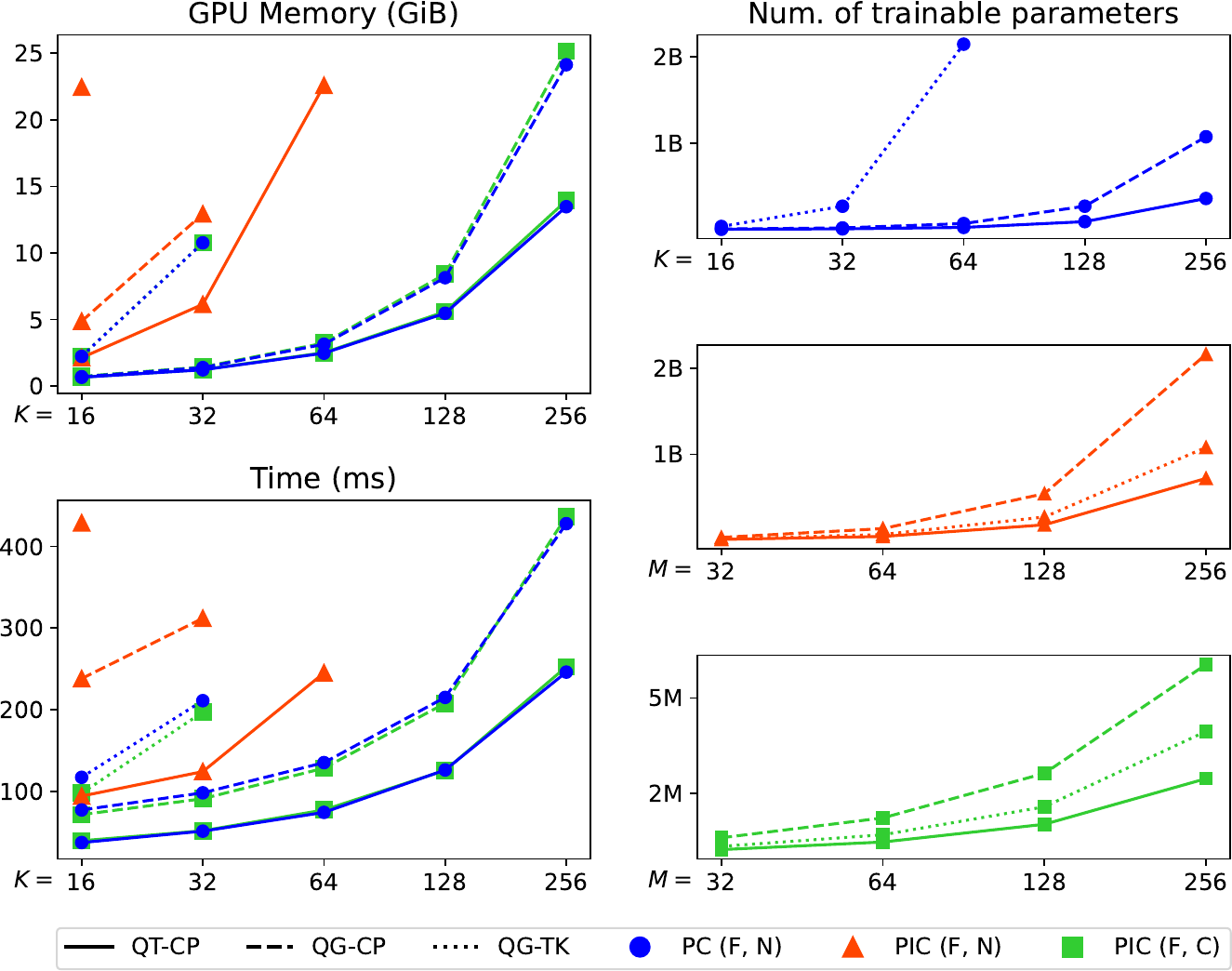}
    \caption{
    \textbf{Learning PICs using functional sharing requires (i) comparable resources as PCs and (ii) up to ~99\% less trainable parameters.}
    We compare the GPU memory (top-left) and time (bottom-left) required to perform an optimization step with PCs (\pcmarker), PICs with functional sharing (\PICsharingmarker), and without (\PICnosharingmarker), while considering three different architectures (\rgqt-\lcp, \rgqg-\lcp, \rgqg-\ltucker).
    To the right, we report the number of trainable parameters for (i) PCs (\pcmarker) at different $K$, and (ii) for PICs (\PICnosharingmarker, \PICsharingmarker) at different MLP sizes $M$.
    The isolated $\PICnosharingmarker$ nodes refer to refer to PIC (F, N) with QG-TK which we could only run at $K \, {=} \, 16$.
    The benchmark is conducted using a batch of 128 RGB images of size 64x64 and Adam \citep{kingma2014adam}.
    Extra details in \cref{app:scaling}.
    }
    \label{fig:bench64x64}
\end{SCfigure}

In our experiments, we first benchmark the effectiveness of functional sharing for scaling the training of PICs via numerical quadrature, comparing it with standard PCs and PICs w/o functional sharing \citep{gala24pic}.
Then, following prior work \citep{dang2022sparse, liu2023scaling, liu2023understanding, gala24pic}, we compare QPCs and PCs as distribution estimators on several image datasets.
We always train using the trapezoidal integration rule.
We use an NVIDIA A100 40GB throughout our experiments.
Our code is available at \href{https://github.com/gengala/ten-pics}{github.com/gengala/ten-pics}.

Thanks to our pipeline, we can now use two recently introduced RGs tailored for image data which deliver architectures that scale better
than those built out of classical RGs \citep{poon2011sum, peharz2020random, liu2021tractable}: \textit{quad-trees} (QTs), tree-shaped RGs, and \textit{quad-graphs} (QGs), DAG-shaped RGs \citep{loconte2024relationship}.
These are perfectly balanced RGs, and therefore applying \cref{alg:rg2pic} over them would deliver balanced PIC structures amenable to depth-wise C-sharing of integral units.
We report full details about QTs and QGs in \cref{app:region-graphs}.
We denote a tensorized architecture as [RG]-[sum-product layer]-[$K$], e.g. \rgqt-\lcp-16, which can be trained as a standard PC or materialized as QPC from a PIC.
We treat pixels as categorical variables, and, as such, our architectures model probability mass functions.

\paragraph{Scaling PICs.}
For each model type, $\{\text{PC}, \text{PIC}\}$, we specify a pair $(\cdot, \cdot)$ where the first (resp. second) argument specifies the sharing technique, $\{\text{F, C, N}\}$, for the input (resp. inner) layers/groups, where N stands for \textit{no sharing}.
In \cref{fig:bench64x64}, we report the time and GPU memory required to perform an Adam \citep{kingma2014adam} optimization step
using PCs (\pcmarker), and PICs with (\PICsharingmarker) and without (\PICnosharingmarker) functional C-sharing over the integral unit groups.
We note that PICs using functional sharing (\PICsharingmarker) proves very effective for scaling, requiring comparable resources as standard PCs (\pcmarker), while those who do not (\PICnosharingmarker)---like prior work \citep{gala24pic}---are orders of magnitude slower and quickly go Out-Of-Memory (OOM) for $K \, {>} \, 64$.
Remarkably, some QG-TK configurations of PICs (\PICsharingmarker), see \cref{tab:bench64x64}, require even less GPU memory than PCs, and this is because of the significant difference in the number of trainable parameters, since copies of these have to be stored by Adam during optimization.
In fact, the number of parameters for PCs and PICs w/o functional sharing (\pcmarker, \PICnosharingmarker) is in the order of hundreds of millions (hitting 2B+), while PICs with functional sharing (\PICsharingmarker) scale much more gracefully, hitting only 6M+ parameters.
We emphasize that the number of trainable parameters of PICs is independent of the $K$ at which we materialize,
but only dependent on the size of the MLPs $M$ we use to parameterize them, which can also be thought as the cost of evaluating PIC functions.
We report more (tabular) details in \cref{app:scaling}.

\begin{figure}
\hfill
{\footnotesize
    \setlength{\tabcolsep}{2.5pt}
    \begin{tabular}{cccccc|cccc}
        \toprule
        & QPC & PC & Sp-PC & HCLT & RAT & IDF & BitS & BBans & McB \\
        \midrule
        \mnist      & \textbf{1.11} & 1.17          & 1.14 & 1.21  & 1.67 & 1.90 & 1.27 & 1.39 & 1.98 \\
        \textsc{f-mnist}    & \textbf{3.16} & 3.32          & 3.27 & 3.34  & 4.29 & 3.47 & 3.28 & 3.66 & 3.72 \\
        \textsc{emn-mn}     & 1.55          & 1.64          & \textbf{1.52} & 1.70  & 2.56 & 2.07 & 1.88 & 2.04 & 2.19 \\
        \textsc{emn-le}     & \textbf{1.54} & 1.62          & 1.58 & 1.75  & 2.73 & 1.95 & 1.84 & 2.26 & 3.12 \\
        \textsc{emn-ba}     & \textbf{1.59} & 1.66          & 1.60 & 1.78  & 2.78 & 2.15 & 1.96 & 2.23 & 2.88 \\
        \textsc{emn-by}     & 1.53          & \textbf{1.47} & 1.54 & 1.73  & 2.72 & 1.98 & 1.87 & 2.23 & 3.14 \\
        \bottomrule
    \end{tabular}
}
\hfill
\includegraphics[align=c,scale=0.30]{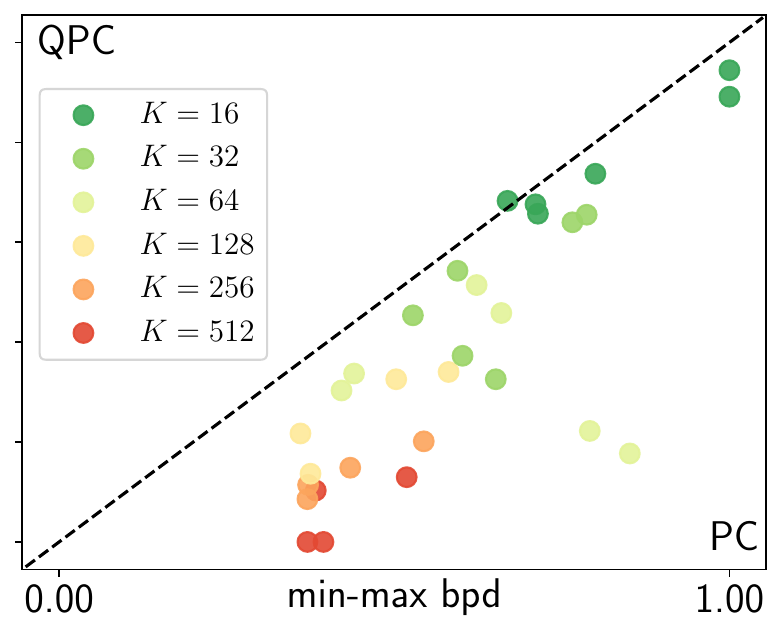}
\hfill\null
\caption{
\textbf{QPCs improve over PCs and other DGM baselines}
in terms of test-set bpd for the \textsc{mnist}-family datasets.
We compare against SparsePC \citep{dang2022sparse}, HCLT \citep{liu2021tractable}, RAT-SPN \citep{peharz2020random}, IDF \citep{hoogeboom2019integer}, BitSwap \citep{kingma2019bit}, BBans \citep{townsend2018practical} and McBits \citep{ruan2021improving}.
HCLT results are taken from \citep{gala24pic}.
Columns QPC and PC report results from this paper, with \rgqg-CP-512 being the best performing architecture for both.
Scatter plot (right): bpd for QPCs (y-axis) and PCs (x-axis) paired by architecture and min-max normalized for the \textsc{mnist} and \textsc{f-mnist} datasets.
A point below the diagonal is a win for QPCs.
}
\label{fig:mnist-results}
\end{figure}

\paragraph{Distribution estimation.}
Following prior work \citep{dang2022sparse, liu2023scaling, liu2023understanding, gala24pic}, we extensively test QPCs and PCs as distribution estimators on standard image datasets.
Our full results are in \cref{app:distribution-estimation}, while we only report here the bits-per-dimension (bpd) of the best performing models, which \textit{always belong to a \rgqg-\lcp architecture, reflecting the additional expressiveness of DAG-shaped RGs.}
Our full results also highlights how the the more expressive yet expensive Tucker layers we introduced for PICs deliver the best performance for small $K$, but are hard to scale.
All QPCs are materialized from PICs applying F-sharing over input units and C-sharing over groups of integral units, i.e.\ PIC (F, C).
We begin with the \mnist-family, which includes 6 datasets of gray-scale 28x28 images: \mnist \citep{mnist}, \fmnist \citep{xiao2017fashion}, and EMNIST with its 4 splits \citep{cohen2017emnist}.
\cref{fig:mnist-results} shows that QPCs generally perform best, improving over standard PCs (5/6), complex heuristic-based PC learning schemes as pruning-and-growing (5/6) \citep{dang2022sparse}, and some deep generative models (DGMs) (6/6).
%


Then, we move to larger RGB image datasets as CIFAR \citep{krizhevsky2009learning}, ImageNet32, ImageNet64 \citep{deng2009ImageNet}, and CelebA \citep{liu2015faceattributes}.
To compare against prior work \citep{liu2023scaling, liu2023understanding}, we have to preprocess the datasets using the YCoCg transform, a lossy color-coding that consistently improves performance for PCs when applied to RGB images.\footnote{The use of the lossy YCoCg transform is undocumented in \citep{liu2023scaling, liu2023understanding} but confirmed via personal communication with the authors.
Note that columns in \cref{table:rgb} using it ($\dagger$) are not directly comparable to the rest.
}
We also report results over datasets preprocessed with the lossless YCoCg-R transform \citep{malvar2003ycocg}, effectively doubling the number of datasets.
We report details about these transforms in \cref{app:ycocg}.
From \cref{table:rgb}, we see that QPCs prove again very competitive, consistently outperforming standard PCs commonly trained with Adam, and the best performing PC from the literature, HCLT, which is trained via EM schemes and patch-wise methods \citep{liu2023scaling}.
Furthermore, QPCs are close to PCs trained via latent variable distillation (LVD, LVD-PG in \cref{table:rgb}) \citep{liu2023scaling, liu2023understanding},
a framework that requires extra supervision over their latent spaces by distilling information from existing deep generative models (DGMs).
This technique requires pre-trained DGMs, several heuristics, and a final fine-tuning stage via EM or SGD, while PIC training method is instead end-to-end and self-contained.

\begin{SCtable}
    \caption{
    \textbf{QPCs improve over PCs.}
    We mark results with * for YCoCg-R and $\dagger$ for YCoCg.
    \rgqg-\lcp-512 (resp.\ \rgqg-\lcp-256) is the best performing architecture for QPCs and PCs on CIFAR \& ImgNet32 (resp.\ ImgNet64 \& CelebA).
    }
    \label{table:rgb}
{\footnotesize
    \setlength{\tabcolsep}{1.5pt}
    \begin{tabular}{ccc|ccc|cccc}
        \toprule
         & $\text{QPC}^*$ & $\text{PC}^*$ & $\text{QPC}^{\dagger}$ & $\text{PC}^{\dagger}$ & $\text{HCLT}^{\dagger}$ & $\text{LVD}^{\dagger}$ & $\text{LVD-PG}^{\dagger}$ \\
        \midrule
        CIFAR         & \textbf{5.09} & 5.50 & \textbf{4.48} & 4.85 & 4.61 & 4.37 & 3.87 \\
        ImgNet32      & \textbf{5.08} & 5.25 & \textbf{4.46} & 4.63 & 4.82 & 4.38 & 4.06 \\
        ImgNet64      & \textbf{5.05} & 5.22 & \textbf{4.42} & 4.59 & 4.67 & 4.12 & 3.80 \\
        CelebA        & \textbf{4.73} & 4.78 & \textbf{4.11} & 4.16 & - & - & - \\
        \bottomrule
    \end{tabular}
}
\end{SCtable}

\section{Discussion \& Conclusion}
\label{sec:discussion}

%
%
%
With this work, we systematized the construction of PICs, extending them to DAG-like structures (\cref{sec:rg2pic}), tensorizing (\cref{sec:pic2qpc}), and scaling their training with functional sharing (\cref{sec:functional sharing}).
In our experiments (\cref{sec:exp}), we showed how this pipeline is remarkably effective when the tractable approximations of PICs, QPCs, are used as distribution estimators.
%
%
%
%
%
This in turn becomes a new and effective tool to learn PCs at scale.
In fact, prior work has shown that naively training large PCs via EM or gradient-ascent is challenging,
and that PC performance plateau as their size increases \citep{dang2022sparse, correia2020joints, liu2023scaling, liu2023understanding}.
Our contributions go beyond these limitations, while offering a simple, principled and fully-differentiable pipeline that delivers performance that rival more sophisticated alternatives \citep{dang2022sparse,liu2023scaling} (\cref{sec:exp}).
We conjecture that this happens as training PCs via PICs drastically reduces the search space while allowing (i) smoother training dynamics and (ii) the materialization of arbitrarily large tractable models.

The development of tractable models is an important task in machine learning as they provide many inference routines, and can be used in many down-stream applications such as tabular data modeling \citep{correia2020joints}, generative modeling \citep{peharz2020einets}, lossless compression \citep{liu2022lossless}, genetics \citep{dang2022tractable}, knowledge-graphs \citep{loconte2023turn}, constrained text generation \citep{zhang2023tractable}, and more.
Our work has also certain parallels with tensor networks and (quasi-)tensor decompositions \citep{tucker64extension, kolda2006multilinear, kim2007nntucker, cichocki2009nnmf, townsend2015continuous}, as \citep{loconte2024relationship}
recently showed how hierarchical tensor decompositions can be represented using the language of tensorized circuits.
Furthermore, we note that the recent non-monotonic PCs \citep{loconte2024subtractive} (i.e., PCs with negative sum parameters) can also be thought as the result of a quadrature process from PICs whose function can return negative values.

Our work does not come without limitations.
Although we showed that training PICs with function sharing requires comparable resources as standard PCs, traditional continuous LV models as VAEs, flows and diffusion models are more scalable.
Also, sampling from PICs is currently not possible, as we cannot perform differentiable sampling from our (multi-headed) MLPs.
Future work may include the investigation of more efficient ways of training PICs, possibly using techniques as LVD or variational inference \citep{kingma2013auto} to directly maximize PIC lower-bounds, requiring numerical quadrature only as fine-tuning step to distill a performant tractable model.
We believe our work will foster new research in the field of generative modeling, and specifically in the realm of tractable models.

\section*{Author Contributions}
GG led the project, proposed neural functional sharing, and ran all experiments.
GG and AV devised the original idea of a pipeline to build PICs, and leverage tensorized folded circuits for training them.
AV and EQ equally supervised all the phases of the project.
CdC supervises the project, and critically read the manuscript and provided feedback.

\section*{Acknowledgments}
The Eindhoven University of Technology authors thank the support from the Eindhoven Artificial Intelligence Systems Institute and the Department of Mathematics and Computer Science of TU Eindhoven.
TU/e authors also thank the support of EU European Defence Fund Project KOIOS (EDF-2021-DIGIT-R-FL-KOIOS).
AV was supported by the "UNREAL: Unified Reasoning Layer for Trustworthy ML" project (EP/Y023838/1) selected by the ERC and funded by UKRI EPSRC.
We thank Lorenzo Loconte for insightful discussions about (quasi-)tensor decompositions.

\bibliography{main.bib}

\begin{thebibliography}{10}

\bibitem{butz2020sum}
Cory Butz, Jhonatan~S Oliveira, and Robert Peharz.
\newblock Sum-product network decompilation.
\newblock In {\em International Conference on Probabilistic Graphical Models}, pages 53--64. PMLR, 2020.

\bibitem{carroll1970indscal}
J.~Douglas Carroll and Jih-Jie Chang.
\newblock Analysis of individual differences in multidimensional scaling via an n-way generalization of “eckart-young” decomposition.
\newblock {\em Psychometrika}, 35:283--319, 1970.

\bibitem{choi2011learning}
Myung~Jin Choi, Vincent Y.~F. Tan, Animashree Anandkumar, and Alan~S Willsky.
\newblock Learning latent tree graphical models.
\newblock {\em Journal of Machine Learning Research}, 12(49):1771--1812, 2011.

\bibitem{choiprobabilistic}
YooJung Choi, Antonio Vergari, and Guy Van~den Broeck.
\newblock Probabilistic circuits: {A} unifying framework for tractable probabilistic models.
\newblock Technical report, UCLA, 2020.

\bibitem{chow1968approximating}
C.~Chow and C.~Liu.
\newblock Approximating discrete probability distributions with dependence trees.
\newblock {\em IEEE Transactions on Information Theory}, 14(3):462--467, 1968.

\bibitem{cichocki2009nnmf}
Andrzej Cichocki and Anh~Huy Phan.
\newblock Fast local algorithms for large scale nonnegative matrix and tensor factorizations.
\newblock {\em {IEICE} Trans. Fundam. Electron. Commun. Comput. Sci.}, 92-A(3):708--721, 2009.

\bibitem{cohen2017emnist}
Gregory Cohen, Saeed Afshar, Jonathan Tapson, and André van Schaik.
\newblock {EMNIST}: Extending {MNIST} to handwritten letters.
\newblock In {\em IJCNN 2017}, pages 2921--2926, 2017.

\bibitem{correia2020joints}
Alvaro Correia, Robert Peharz, and Cassio~P de~Campos.
\newblock Joints in random forests.
\newblock In {\em Advances in Neural Information Processing Systems}, volume~33, pages 11404--11415, 2020.

\bibitem{correia2023continuous}
Alvaro H.~C. Correia, Gennaro Gala, Erik Quaeghebeur, Cassio de~Campos, and Robert Peharz.
\newblock Continuous mixtures of tractable probabilistic models.
\newblock {\em Proceedings of the AAAI Conference on Artificial Intelligence}, 37(6):7244--7252, 2023.

\bibitem{dang2022sparse}
Meihua Dang, Anji Liu, and Guy~Van den Broeck.
\newblock Sparse probabilistic circuits via pruning and growing.
\newblock In {\em NeurIPS 2022}, volume~35 of {\em Advances in Neural Information Processing Systems}, 2022.

\bibitem{dang2022tractable}
Meihua Dang, Anji Liu, Xinzhu Wei, Sriram Sankararaman, and Guy Van~den Broeck.
\newblock Tractable and expressive generative models of genetic variation data.
\newblock In {\em Research in Computational Molecular Biology}, pages 356--357, 2022.

\bibitem{darwiche2009modeling}
Adnan Darwiche.
\newblock {\em Modeling and reasoning with Bayesian networks}.
\newblock Cambridge University Press, 2009.

\bibitem{davis2007methods}
Philip~J. Davis and Philip Rabinowitz.
\newblock {\em Methods of numerical integration}.
\newblock Academic Press, 1984.

\bibitem{deng2009ImageNet}
Jia Deng, Wei Dong, Richard Socher, Li-Jia Li, Kai Li, and Li~Fei-Fei.
\newblock Imagenet: A large-scale hierarchical image database.
\newblock In {\em 2009 IEEE conference on computer vision and pattern recognition}, pages 248--255. Ieee, 2009.

\bibitem{dennis2012learning}
Aaron~W. Dennis and Dan Ventura.
\newblock Learning the architecture of sum-product networks using clustering on variables.
\newblock In {\em Advances in Neural Information Processing Systems 25 (NeurIPS)}, pages 2033--2041. Curran Associates, Inc., 2012.

\bibitem{di2021random}
Nicola {Di Mauro}, Gennaro Gala, Marco Iannotta, and Teresa Maria~Altomare Basile.
\newblock Random probabilistic circuits.
\newblock In {\em 37th Conference on Uncertainty in Artificial Intelligence (UAI)}, volume 161, pages 1682--1691. PMLR, 2021.

\bibitem{fubini1907sugli}
Guido Fubini.
\newblock Sugli integrali multipli.
\newblock {\em Rend. Acc. Naz. Lincei}, 16:608--614, 1907.

\bibitem{gala24pic}
Gennaro Gala, Cassio de~Campos, Robert Peharz, Antonio Vergari, and Erik Quaeghebeur.
\newblock Probabilistic integral circuits.
\newblock In {\em Proceedings of The 27th International Conference on Artificial Intelligence and Statistics}, volume 238 of {\em Proceedings of Machine Learning Research}, pages 2143--2151. PMLR, 02--04 May 2024.

\bibitem{gens2013learning}
Robert Gens and Pedro~M. Domingos.
\newblock Learning the structure of sum-product networks.
\newblock In {\em International Conference on Machine Learning}, 2013.

\bibitem{goodfellow2014generative}
Ian Goodfellow, Jean Pouget-Abadie, Mehdi Mirza, Bing Xu, David Warde-Farley, Sherjil Ozair, Aaron Courville, and Yoshua Bengio.
\newblock Generative adversarial nets.
\newblock In {\em NIPS 2014}, volume~27 of {\em Advances in Neural Information Processing Systems}, 2014.

\bibitem{hoogeboom2019integer}
Emiel Hoogeboom, Jorn Peters, Rianne van~den Berg, and Max Welling.
\newblock Integer discrete flows and lossless compression.
\newblock In {\em NeurIPS 2019}, volume~32 of {\em Advances in Neural Information Processing Systems}, 2019.

\bibitem{kim2007nntucker}
Yong{-}Deok Kim and Seungjin Choi.
\newblock Nonnegative tucker decomposition.
\newblock In {\em {CVPR}}. {IEEE} Computer Society, 2007.

\bibitem{kingma2014adam}
Diederik~P. Kingma and Jimmy Ba.
\newblock Adam: {A} method for stochastic optimization.
\newblock In {\em ICLR 2015}, 2015.

\bibitem{kingma2013auto}
Diederik~P. Kingma and Max Welling.
\newblock Auto-encoding variational {Bayes}.
\newblock In {\em ICLR 2014}, 2014.

\bibitem{kingma2019bit}
Friso Kingma, Pieter Abbeel, and Jonathan Ho.
\newblock Bit-swap: Recursive bits-back coding for lossless compression with hierarchical latent variables.
\newblock In {\em Proceedings of the 36th International Conference on Machine Learning}, volume~97 of {\em Proceedings of Machine Learning Research}, pages 3408--3417, 2019.

\bibitem{kolda2006multilinear}
Tamara~G. Kolda.
\newblock Multilinear operators for higher-order decompositions.
\newblock Technical report, {Sandia National Laboratories}, 2006.

\bibitem{koller2009probabilistic}
Daphne Koller and Nir Friedman.
\newblock {\em Probabilistic graphical models: principles and techniques}.
\newblock MIT press, 2009.

\bibitem{krizhevsky2009learning}
Alex Krizhevsky.
\newblock Learning multiple layers of features from tiny images, 2009.

\bibitem{mnist}
Yann LeCun, Corinna Cortes, and Christopher J.~C. Burges.
\newblock The {MNIST} database of handwritten digits, 2010.

\bibitem{liu2022lossless}
Anji Liu, Stephan Mandt, and Guy Van~den Broeck.
\newblock Lossless compression with probabilistic circuits.
\newblock In {\em ICLR 2022}, 2022.

\bibitem{liu2021tractable}
Anji Liu and Guy Van~den Broeck.
\newblock Tractable regularization of probabilistic circuits.
\newblock In {\em NeurIPS 2021}, volume~34 of {\em Advances in Neural Information Processing Systems}, pages 3558--3570, 2021.

\bibitem{liu2023scaling}
Anji Liu, Honghua Zhang, and Guy Van~den Broeck.
\newblock Scaling up probabilistic circuits by latent variable distillation.
\newblock In {\em ICLR 2023}, 2023.

\bibitem{liu2023understanding}
Xuejie Liu, Anji Liu, Guy Van~den Broeck, and Yitao Liang.
\newblock Understanding the distillation process from deep generative models to tractable probabilistic circuits.
\newblock In {\em Proceedings of the 40th International Conference on Machine Learning}, volume 202 of {\em Proceedings of Machine Learning Research}, pages 21825--21838, 2023.

\bibitem{liu2015faceattributes}
Ziwei Liu, Ping Luo, Xiaogang Wang, and Xiaoou Tang.
\newblock Deep learning face attributes in the wild.
\newblock In {\em Proceedings of International Conference on Computer Vision (ICCV)}, December 2015.

\bibitem{loconte2024subtractive}
Lorenzo Loconte, M.~Sladek Aleksanteri, Stefan Mengel, Martin Trapp, Arno Solin, Nicolas Gillis, and Antonio Vergari.
\newblock Subtractive mixture models via squaring: Representation and learning.
\newblock In {\em The Twelfth International Conference on Learning Representations ({ICLR})}, 2024.

\bibitem{loconte2024relationship}
Lorenzo Loconte, Antonio Mari, Gennaro Gala, Robert Peharz, Cassio de~Campos, Erik Quaeghebeur, Gennaro Vessio, and Antonio Vergari.
\newblock What is the relationship between tensor factorizations and circuits (and how can we exploit it)?
\newblock {\em arXiv preprint arXiv:2409.07953}, 2024.

\bibitem{loconte2023turn}
Lorenzo Loconte, Nicola~Di Mauro, Robert Peharz, and Antonio Vergari.
\newblock How to turn your knowledge graph embeddings into generative models via probabilistic circuits.
\newblock In {\em Advances in Neural Information Processing Systems 37 (NeurIPS)}. Curran Associates, Inc., 2023.

\bibitem{loshchilov2017sgdr}
Ilya Loshchilov and Frank Hutter.
\newblock {SGDR}: Stochastic gradient descent with warm restarts.
\newblock In {\em ICLR 2017}, 2017.

\bibitem{malvar2003ycocg}
Henrique Malvar and Gary Sullivan.
\newblock Ycocg-r: A color space with rgb reversibility and low dynamic range.
\newblock {\em ISO/IEC JTC1/SC29/WG11 and ITU-T SG16 Q}, 6, 2003.

\bibitem{molina2018mixed}
Alejandro Molina, Antonio Vergari, Nicola {Di Mauro}, Sriraam Natarajan, Floriana Esposito, and Kristian Kersting.
\newblock Mixed sum-product networks: A deep architecture for hybrid domains.
\newblock In {\em AAAI Conference on Artificial Intelligence}, 2018.

\bibitem{peharz2016latent}
Robert Peharz, Robert Gens, Franz Pernkopf, and Pedro Domingos.
\newblock On the latent variable interpretation in sum-product networks.
\newblock {\em IEEE Transactions on Pattern Analysis and Machine Intelligence}, 39(10):2030--2044, 2017.

\bibitem{peharz2020einets}
Robert Peharz, Steven Lang, Antonio Vergari, Karl Stelzner, Alejandro Molina, Martin Trapp, Guy Van Den~Broeck, Kristian Kersting, and Zoubin Ghahramani.
\newblock Einsum networks: Fast and scalable learning of tractable probabilistic circuits.
\newblock In {\em 37th International Conference on Machine Learning (ICML)}, volume 119 of {\em Proceedings of Machine Learning Research}, pages 7563--7574. PMLR, 2020.

\bibitem{peharz2020random}
Robert Peharz, Antonio Vergari, Karl Stelzner, Alejandro Molina, Xiaoting Shao, Martin Trapp, Kristian Kersting, and Zoubin Ghahramani.
\newblock Random sum-product networks: A simple and effective approach to probabilistic deep learning.
\newblock In {\em Proceedings of The 35th Uncertainty in Artificial Intelligence Conference}, volume 115 of {\em Proceedings of Machine Learning Research}, pages 334--344, 2020.

\bibitem{pipatsrisawat2008new}
Knot Pipatsrisawat and Adnan Darwiche.
\newblock New compilation languages based on structured decomposability.
\newblock In {\em Proceedings of the 23rd National Conference on Artificial Intelligence (AAAI'08)}, volume~1, pages 517--522, 2008.

\bibitem{poon2011sum}
Hoifung Poon and Pedro Domingos.
\newblock Sum-product networks: A new deep architecture.
\newblock In {\em {IEEE} International Conference on Computer Vision Workshops (ICCV Workshops)}, pages 689--690. IEEE, 2011.

\bibitem{ruan2021improving}
Yangjun Ruan, Karen Ullrich, Daniel~S. Severo, James Townsend, Ashish Khisti, Arnaud Doucet, Alireza Makhzani, and Chris Maddison.
\newblock Improving lossless compression rates via monte carlo bits-back coding.
\newblock In {\em Proceedings of the 38th International Conference on Machine Learning}, volume 139 of {\em Proceedings of Machine Learning Research}, pages 9136--9147, 2021.

\bibitem{segal2005learning}
Eran Segal, Dana Pe'er, Aviv Regev, Daphne Koller, Nir Friedman, and Tommi Jaakkola.
\newblock Learning module networks.
\newblock {\em Journal of Machine Learning Research}, 6(4), 2005.

\bibitem{tancik2020fourier}
Matthew Tancik, Pratul Srinivasan, Ben Mildenhall, Sara Fridovich-Keil, Nithin Raghavan, Utkarsh Singhal, Ravi Ramamoorthi, Jonathan Barron, and Ren Ng.
\newblock Fourier features let networks learn high frequency functions in low dimensional domains.
\newblock In {\em NeurIPS 2020}, volume~33 of {\em Advances in Neural Information Processing Systems}, pages 7537--7547, 2020.

\bibitem{townsend2015continuous}
Alex Townsend and Lloyd~N Trefethen.
\newblock Continuous analogues of matrix factorizations.
\newblock {\em Proceedings of the Royal Society A: Mathematical, Physical and Engineering Sciences}, 471(2173):20140585, 2015.

\bibitem{townsend2018practical}
James Townsend, Thomas Bird, and David Barber.
\newblock Practical lossless compression with latent variables using bits back coding.
\newblock In {\em ICLR 2019}, 2019.

\bibitem{tucker64extension}
L.~R. Tucker.
\newblock {T}he extension of factor analysis to three-dimensional matrices.
\newblock In {\em {C}ontributions to mathematical psychology.}, pages 110--127. Holt, Rinehart and Winston, 1964.

\bibitem{vergari2021compositional}
Antonio Vergari, YooJung Choi, Anji Liu, Stefano Teso, and Guy Van~den Broeck.
\newblock A compositional atlas of tractable circuit operations for probabilistic inference.
\newblock In {\em NeurIPS 2021}, volume~36 of {\em Advances in Neural Information Processing Systems}, 2021.

\bibitem{vergari2019tractable}
Antonio Vergari, Nicola Di~Mauro, and Guy Van~den Broeck.
\newblock Tractable probabilistic models: {R}epresentations, algorithms, learning, and applications, 2019.
\newblock Tutorial at the 35th Conference on Uncertainty in Artificial Intelligence (UAI 2019).

\bibitem{vergari2018sum}
Antonio Vergari, Robert Peharz, Nicola Di~Mauro, Alejandro Molina, Kristian Kersting, and Floriana Esposito.
\newblock Sum-product autoencoding: Encoding and decoding representations using sum-product networks.
\newblock {\em Proceedings of the AAAI Conference on Artificial Intelligence}, 32(1), 2018.

\bibitem{xiao2017fashion}
Han {Xiao}, Kashif {Rasul}, and Roland {Vollgraf}.
\newblock {Fashion-MNIST}: a novel image dataset for benchmarking machine learning algorithms.
\newblock {\em arXiv}, 2017.

\bibitem{yang2023diffusion}
Ling Yang, Zhilong Zhang, Yang Song, Shenda Hong, Runsheng Xu, Yue Zhao, Wentao Zhang, Bin Cui, and Ming-Hsuan Yang.
\newblock Diffusion models: A comprehensive survey of methods and applications.
\newblock {\em ACM Computing Surveys}, 56(4):1--39, 2023.

\bibitem{yang2023bayesian}
Yang Yang, Gennaro Gala, and Robert Peharz.
\newblock Bayesian structure scores for probabilistic circuits.
\newblock In {\em Proceedings of The 26th International Conference on Artificial Intelligence and Statistics}, volume 206 of {\em Proceedings of Machine Learning Research}, pages 563--575, 2023.

\bibitem{zhang2023tractable}
Honghua Zhang, Meihua Dang, Nanyun Peng, and Guy~Van den Broeck.
\newblock Tractable control for autoregressive language generation.
\newblock In {\em 40th International Conference on Machine Learning (ICML)}, volume 202 of {\em Proceedings of Machine Learning Research}, pages 40932--40945. {PMLR}, 2023.

\end{thebibliography}
\bibliographystyle{plain}

\clearpage
\newpage
\appendix

\counterwithin{table}{section}
\counterwithin{figure}{section}
\counterwithin{algorithm}{section}
\renewcommand{\thetable}{\thesection.\arabic{table}}
\renewcommand{\thefigure}{\thesection.\arabic{figure}}
\renewcommand{\thealgorithm}{\thesection.\arabic{algorithm}}

\section{Background on Circuits}
\label{app:circuit}


\begin{definition}[Circuit \citep{choiprobabilistic, vergari2021compositional}]
    A circuit $c$ over variables $\X$ is a parameterized computational graph encoding a function $c(\X)$, and comprising three kinds of computational units: input, product, and sum.
    Each product or sum unit $u$ outputs a scalar and receives as inputs the output scalars of other units, denoted with the set $\textsf{in}(u)$.
    Each unit $u$ computes a function $f_u$ defined as: (i) $f_u(\X_u) \, {\rightarrow} \, \sR$ if $u$ is an input unit, where $f_u$ is a function over variables $\X_u \subseteq \X$, called its scope,
    (ii) $\Pi_{i \in \textsf{in}(u)} f_i (\X_i)$ if $u$ is a product unit,
    and (iii) $\Sigma_{i \in \textsf{in}(u)} f_i (\X_i)$ if $u$ is a sum unit, with $w_i \in \sR$ denoting the weighted sum parameters.
    The scope of a product or sum unit is the union of the scopes of its input units.
\end{definition}

\begin{definition}[Probabilistic Circuit]
    A PC over variables $\X$ is a circuit $c$ encoding a (possibly non-normalized) distribution, e.g., a function that is non-negative for all values of $\X$:
    \begin{equation*}
        c(\x) \ge 0, \quad \forall \x \in \mathsf{dom}(\X)
    \end{equation*}
\end{definition}

\begin{definition}[Smoothness]
    A circuit is \textit{smooth} if, for each sum unit $u$, its inputs depend on the same variables: $\forall u_1, u_2 \in \mathsf{in}(u), \X_{u_1} = \X_{u_2}$.
\end{definition}

\begin{definition}[Decomposability]
    A circuit is \textit{decomposable} if the inputs of each product unit $u$ depend on disjoint sets of variables: $\forall u_1, u_2 \in \mathsf{in}(u), \X_{u_1} \neq \X_{u_2}$.
\end{definition}

\begin{definition}[Structured-decomposability \citep{pipatsrisawat2008new, darwiche2009modeling}]
 A circuit is \emph{structured-decomposable} if (i) it is smooth and decomposable, and (2) any pair of product units having the same scope decompose their scope at their input units in the same way.
\end{definition}

Although all tensorized architectures mentioned in this paper are smooth and decomposable, only PCs built from tree RGs are also structured-decomposable, and as such are potentially less expressive because they belong to a restricted class.

\section{Region Graphs}
\label{app:region-graphs}

In \cref{alg:build-quad-graph} we detail the construction of the Quad-Tree (\rgqt) and Quad-Graph (\rgqg) region graphs \citep{loconte2024relationship}.
Specifically, QTs (resp. QGs) are built setting the input flag \textsf{isTree} to True (resp. False).
Intuitively, these RGs recursively split an image into patches, until reaching regions associated to exactly one pixel.
The splitting is performed both horizontally and vertically, and subsequent patches can either be shared, thus yielding a RG that is not a tree (QGs), or not (QTs).

\begin{figure}[H]
\begin{minipage}[t]{0.53\linewidth}
\begin{algorithm}[H]
    \small
    \caption{$\textsf{buildQuadGraph}(H,W,\textsf{isTree})$}
    \label{alg:build-quad-graph}
    \textbf{Input:} Image height $H$, image width $W$, and whether to enforce the output RG to be a tree. \\
    \textbf{Output:} A RG $\gR$ over $H\cdot W$ variables
    \begin{algorithmic}[1]
        \State $\textsf{S}\leftarrow \{\reg_{ij}=\{X_{ij}\}\mid (i, j) \in [H] \times [W]\}$
        \State $\gR \leftarrow$ a RG with leaf regions $\textsf{S}$
        \State $h\leftarrow H$;\ \ $w\leftarrow W$
        \While{$h > 1\lor w > 1$}
            \State $h\leftarrow\lceil h/2 \rceil$;\ \ $w\leftarrow\lceil w/2 \rceil$;\ \ $\textsf{S}' \leftarrow \varnothing$
            \For{$i,j\in [h]\times [w]$}
                \State $\Omega {\leftarrow} (\{2i{-}1, 2i\}{\times}\{2j{-}1, 2j\}){\cap}([H] {\times} [W])$
                \If{$|\Omega| = 1$}
                    \State Let $\reg_{pq}\in\textsf{S}$ s.t. $(p,q)\in\Omega$
                    \State $\textsf{addRegion}(\gR,\reg_{pq})$
                \ElsIf{$|\Omega| = 2$}
                    \State Let $\reg_{pq},\reg_{rs}\in\textsf{S}$ s.t.\\
                    \hspace*{6em} $(p,q),(r,s)\in\Omega,\quad p<r,q<s$
                    \State $\textsf{addPartition}(\gR,\reg_{pq}\cup\reg_{rs},\{\reg_{pq},\reg_{rs}\})$
                \Else $\quad \, \triangleright \, |\Omega| = 4$
                    \If{\textsf{isTree}}
                        $\textsf{mergeTree}(\gR,\Omega,\textsf{S})$
                    \Else
                        \ $\textsf{mergeDAG}(\gR,\Omega,\textsf{S})$
                    \EndIf
                \EndIf
                    \State $\reg_{ij} \leftarrow \bigcup_{(r,s) \in \Omega} \reg_{rs}$ s.t. $\reg_{rs}\in\textsf{S}$
                    \State $\textsf{S}'\leftarrow\textsf{S}'\cup\{\reg_{ij}\}$
            \EndFor
            \State $\textsf{S}\leftarrow \textsf{S}'$
        \EndWhile
        \State \Return $\gR$
    \end{algorithmic}
\end{algorithm}
\end{minipage}
\hfill
\begin{minipage}[t]{0.45\linewidth}
\begin{algorithm}[H]
    \small
    \caption{$\textsf{mergeTree}(\gR,\Omega,\textsf{S})$}
    \label{alg:build-quad-graph-split-tree}
    \textbf{Input:} A RG $\gR$, a set of four coordinates $\Omega$, and a set of regions $\textsf{S}$ \\
    \textbf{Behavior:} It merges the regions indexed by $\Omega$ in $\gR$ by forming a tree structure
    \begin{algorithmic}[1]
        \State Let $\reg_{uv}=\mathbb{Y}_{p+u\:q+v}\in\textsf{S}$ s.t.\\
        \qquad $(p+u,q+v)\in\Omega, \quad u,v\in\{0,1\}$
    \State $\reg \leftarrow \reg_{00} \cup \reg_{01} \cup \reg_{10} \cup \reg_{11}$
    \State $\textsf{addPartition}(\gR, \reg, \{\reg_{00},\reg_{01},\reg_{10},\reg_{11}\})$
    \end{algorithmic}
\end{algorithm}
\vspace{-11pt}
\begin{algorithm}[H]
    \small
    \caption{$\textsf{mergeDAG}(\gR,\Omega,\textsf{S})$}
    \label{alg:build-quad-graph-split-dag}
    \textbf{Input:} A RG $\gR$, a set of four coordinates $\Omega$, and a set of regions $\textsf{S}$ \\
    \textbf{Behavior:} It merges the regions indexed by $\Omega$ in $\gR$ by forming a DAG structure
    \begin{algorithmic}[1]
        \State Let $\reg_{uv}=\mathbb{Y}_{p+u\:q+v}\in\textsf{S}$ s.t.\\
        \qquad $(p+u,q+v)\in\Omega, \quad u,v\in\{0,1\}$
        \State $\reg\leftarrow \reg_{00}\cup\reg_{01}\cup \reg_{10}\cup\reg_{11}$
        \State $\textsf{addPartition}(\gR,\reg,\{\reg_{00}\cup\reg_{01},\reg_{10}\cup\reg_{11}\})$
        \State $\textsf{addPartition}(\gR,\reg,\{\reg_{00}\cup\reg_{10},\reg_{01}\cup\reg_{11}\})$
        \State $\textsf{addPartition}(\gR,\reg_{00}\cup\reg_{01},\{\reg_{00},\reg_{01}\})$
        \State $\textsf{addPartition}(\gR,\reg_{10}\cup\reg_{11},\{\reg_{10},\reg_{11}\})$
        \State $\textsf{addPartition}(\gR,\reg_{00}\cup\reg_{10},\{\reg_{00},\reg_{10}\})$
        \State $\textsf{addPartition}(\gR,\reg_{01}\cup\reg_{11},\{\reg_{01},\reg_{11}\})$
    \end{algorithmic}
\end{algorithm}
\end{minipage}
\end{figure}

\clearpage
\section{Implementation details}
\label{app:implementation}

\subsection{Multi-headed MLP details}
\label{app:mlp-details}

A multi-headed MLP of size $M$ parameterizing a group $\gamma \, {=} \, \{u_i\}_{i=1}^{\smash{N}}$ of $N$ PIC units with functions of the form $\sR^I \, {\rightarrow} \, \sR^O$ consists of:
\begin{enumerate}
    \item A Fourier-Features Layer (details below), i.e. a non-linear mapping $\sR^I \rightarrow \sR^M$;
    \item Two linear layers followed by hyperbolic tangent as activation function, i.e.\ two consecutive non-linear mappings $\sR^M \rightarrow \sR^M$;
    \item $N$ heads with Softplus non-linearity, i.e.\ $N$ different non-linear mapping $\sR^M \, {\rightarrow} \, \sR^O$.
\end{enumerate}

Note that, if $\gamma$ is a group of \textsf{CP} (resp.\ \textsf{Tucker}) integral units, then $I=2$ (resp.\ $I=3$), while the output dimension $O$ is always equal to 1.
Instead, if the group $\gamma$ is a group of input units, the input dimension $I$ is always equal to 1, while the output dimension $O$ is equal to the number of required parameters of the specific distribution, e.g.\ $O \, {=} \, 2$ for Gaussians.

Such multi-headed MLP is implemented using grouped 1D convolutions, which allow a one-shot materialization of all the layer parameters associated to the group.
We found that initializing all the heads to be equal improves convergence.

\paragraph{Fourier Feature Layer.}
Fourier Feature Layers (FFLs) are an important ingredient for the multi-headed MLPs.
FFLs \citep{tancik2020fourier} enable MLPs to learn high-frequency functions in low-dimensional problem domains and are usually used as first layers of coordinate-based MLPs.
FFLs transform input $\rvz \in \sR^I$ to
\[
\textsf{FFL}(\rvz): \sR^I \rightarrow \sR^M := [ \cos (2 \pi \rvf_1^{\intercal} \rvz),  \sin (2 \pi \rvf_1^{\intercal} \rvz), \dots, \cos (2 \pi \rvf_{M/2}^{\intercal} \rvz),  \sin (2 \pi \rvf_{M/2}^{\intercal} \rvz)],
\]
where $M$ is a hyper-parameter and vectors $\rvf_i \in \sR^I$ are non-learnable, randomly initialized parameters.
FFLs have two main benefits: (i) They allow learning more expressive functions by avoiding over-smoothing behaviors, and (ii) they reduce the total number of trainable parameters when used instead of conventional linear layers as the initial layers in MLPs.

\subsection{Training details}
We train both PICs and PCs using the same training setup.
Specifically, for each dataset, we perform a training cycle of $T$ optimization steps, after which we perform a validation step and stop training if the validation log-likelihood did not improve by $\delta$ nats after 5 training cycles.
Using $\delta \, {>} \, 0$ can avoid long trainings with negligible improvements.
We report these common training hyper-parameters in \cref{tab:hyperparam}.
We use Adam \citep{kingma2014adam} and a batch size of 256 for all experiments.

\paragraph{PIC training.}
After some preliminary runs, we found that a learning rate of $5 \times 10^{-3}$ worked best, which we annealed towards $10^{-4}$ using cosine annealing with warm restarts across 500 optimization steps \citep{loshchilov2017sgdr}.
We also apply weight decay with $\lambda = 0.01$.

\paragraph{PC training.}
After some preliminary runs, we found that a constant learning rate of 0.01 worked best for all PC models, and for all datasets.
We keep the PC parameters unnormalized, and, as such, we clamp them to a small positive value ($10^{-19}$) after each Adam update to keep them non-negative, and subtract the log normalization constant to normalize the log-likelihoods.

\begin{table}[H]
\caption{Common training hyper-parameters for PICs and PCs.}
\label{tab:hyperparam}
\centering
\begin{tabular}{cccc}
 \hline
 dataset            & max num epochs & $T$ & $\delta$ \\
 \hline
 \mnist-family excl.\ \textsc{emnist-by}    & 200   & 250   & 0     \\
 \textsc{emnist-by}                         & 100   & 1000  & 0     \\
 CIFAR                                      & 200   & 250   & 0     \\
 ImageNet32                                 & 50    & 2000  & 10    \\
 ImageNet64                                 & 50    & 2000  & 30    \\
 CelebA                                     & 200   & 750   & 10    \\
 \hline
\end{tabular}
\end{table}

\clearpage
\subsection{YCoCg color-coding transforms}
\label{app:ycocg}

\begin{figure}[H]
\includegraphics[width=\textwidth]{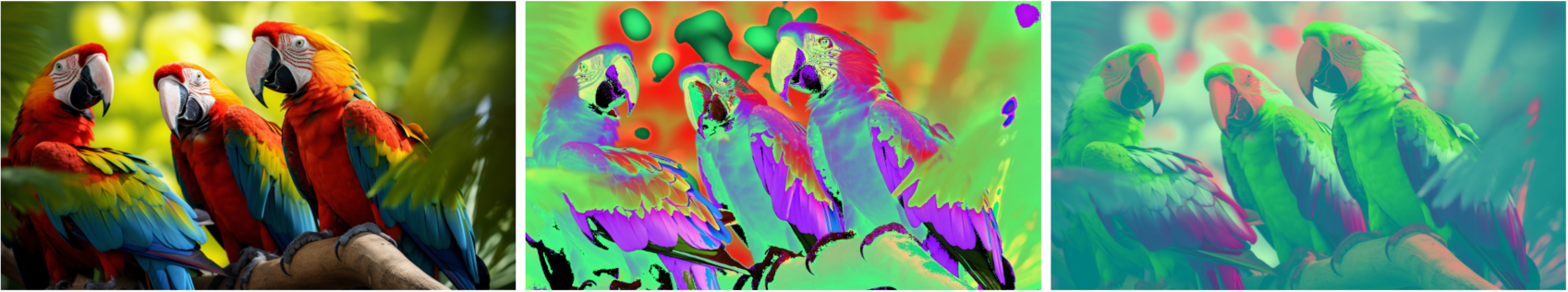}
\caption{\textbf{Visual difference of YCoCg-R and YCoCg.}
Given the RGB image to the left, we show the application of YCoCg-R in the middle and that of YCoCg to the right.
}
\label{fig:parrots-ycc}
\end{figure}

In \cref{fig:ycc-lossless} and \cref{fig:ycc-lossy} we provide pytorch code for the lossless and lossy versions of the YCoCg transform that we used in our experiments (\cref{sec:exp}).
In \cref{fig:ycc-use-case}, we show how to apply them and that the lossy version is on average off less than a bit.
Finally, in \cref{fig:parrots-ycc} we show the significant visual difference of the two transforms when applied to an RGB image.

\begin{figure}[H]
\begin{minipage}{0.48\linewidth}
\small{
    \begin{pythoncode}
def rgb2ycc_lossless(
    rgb_img: torch.Tensor
):
    assert rgb_img.size(-1) == 3

    def forward_lift(x, y):
        diff = (y - x) % 256
        average = (x + (diff >> 1)) % 256
        return average, diff

    red = rgb_img[..., 0]
    green = rgb_img[..., 1]
    blue = rgb_img[..., 2]

    temp, co = forward_lift(red, blue)
    y, cg = forward_lift(green, temp)
    ycc_img = torch.stack(
        [y, co, cg], dim=-1
    )
    return ycc_img
    \end{pythoncode}
}
\end{minipage}
    \hfill
\begin{minipage}{0.48\linewidth}
\small{
    \begin{pythoncode}
def ycc2rgb_lossless(
    ycc_img: torch.Tensor
):
    assert ycc_img.size(-1) == 3

    def reverse_lift(average, diff):
        x = (average - (diff >> 1)) % 256
        y = (x + diff) % 256
        return x, y

    y = ycc_img[..., 0]
    co = ycc_img[..., 1]
    cg = ycc_img[..., 2]

    green, temp = reverse_lift(y, cg)
    red, blue = reverse_lift(temp, co)
    rgb_img = torch.stack(
        [red, green, blue], dim=-1
    )
    return rgb_img
    \end{pythoncode}
}
\end{minipage}
\vspace{3mm}
\caption{\textbf{Lossless YCoCg transform (aka YCoCg-R \citep{malvar2003ycocg}).}
We attach pytorch code for the $\textsf{RGB} \rightarrow \textsf{YCoCg}$ direction (left) and $\textsf{YCoCg} \rightarrow \textsf{RGB}$ direction (right), where one is the inverse of the other.
The input to both functions is a tensor with discrete values in $[0, 255]$, so their output.
}
\label{fig:ycc-lossless}
\end{figure}

\begin{figure}[H]
\begin{minipage}{0.48\linewidth}
\small{
    \begin{pythoncode}
def rgb2ycc_lossy(
    rgb_img: torch.Tensor
):
    assert rgb_img.size(-1) == 3

    deq_img = (rgb_img / 127.5) - 1
    red = (deq_img[..., 0] + 1) / 2
    green = (deq_img[..., 1] + 1) / 2
    blue = (deq_img[..., 2] + 1) / 2

    co = red - blue
    tmp = blue + co / 2
    cg = green - tmp
    y = tmp + cg / 2
    y = y * 2 - 1

    transformed_img = torch.stack(
        [y, co, cg], dim=-1
    )
    ycc_img = torch.floor(
        ((transformed_img + 1) / 2) * 256
    ).long().clip(0, 255)
    return ycc_img
    \end{pythoncode}
}
\end{minipage}
    \hfill
\begin{minipage}{0.48\linewidth}
\small{
    \begin{pythoncode}
def ycc2rgb_lossy(
    ycc_img: torch.Tensor
):
    assert ycc_img.size(-1) == 3

    deq_img = (ycc_img / 127.5) - 1
    y = deq_img[..., 0]
    co = deq_img[..., 1]
    cg = deq_img[..., 2]

    y = (y + 1) / 2
    tmp = y - cg / 2
    green = cg + tmp
    blue = tmp - co / 2
    red = blue + co

    transformed_img = torch.stack(
        [red, green, blue], dim=-1
    )
    rgb_img = torch.floor(
        (transformed_img * 255)
    ).long().clip(0, 255)
    return rgb_img
    \end{pythoncode}
}
\end{minipage}
\vspace{3mm}
\caption{\textbf{Lossy YCoCg transform.}
We attach pytorch code for the $\textsf{RGB} \rightarrow \textsf{YCoCg}$ direction (left) and $\textsf{YCoCg} \rightarrow \textsf{RGB}$ direction (right).
The two functions do \textit{not} represent a bijection.
The input to both functions is a tensor with discrete values in $[0, 255]$, so their output.
}
\label{fig:ycc-lossy}
\end{figure}

\begin{figure}[H]
\small{
    \begin{pythoncode}
batch_size = 100
img_size = 32
rgb_batch = torch.randint(256, (batch_size, img_size * img_size, 3))

ycc_batch_lossless = rgb2ycc_lossless(rgb_batch)
recon_rgb_batch_lossless = ycc2rgb_lossless(ycc_batch_lossless)
print((recon_rgb_batch_lossless == rgb_batch).all())  # True

ycc_batch_lossy = rgb2ycc_lossy(rgb_batch)
recon_rgb_batch_lossy = ycc2rgb_lossy(ycc_batch_lossy)
print((rgb_batch - recon_rgb_batch_lossy).abs().float().mean())  # around 0.66
    \end{pythoncode}
}
\caption{\textbf{Application of the YCoCg transforms.}
We show that YCoCg-R is indeed a bijection, and that the lossy YCoCg is on average off less than a bit.
}
\label{fig:ycc-use-case}
\end{figure}

\section{Additional results}

\subsection{Scaling experiments}
\label{app:scaling}

\begin{figure*}[!ht]
	\hfill
    \includegraphics[align=c, scale=0.35]{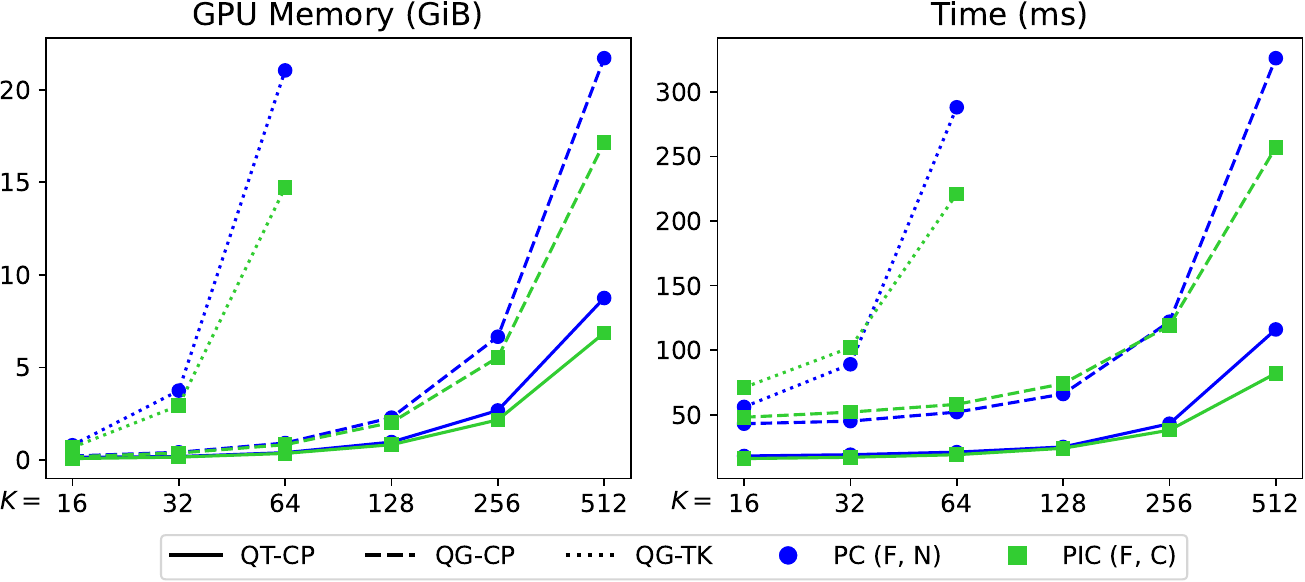}
    \hfill
    {\scriptsize
        \begin{tabular}{crrr}
             &   \rgqt-\lcp &  \rgqg-\lcp & \rgqg-\ltucker     \\
             \toprule
             PIC&          1.1M    &  2.2M     & 1.8M      \\
             \toprule
             $K{=}16$&      270K    &  800K     & 6M      \\
             $K{=}32$&      1M      &  3M       & 51M     \\
             $K{=}64$&      4M      &  13M      & 408M      \\
             $K{=}128$&     17M     &  51.1M    & -       \\
             $K{=}256$&     69M     &  204M     & -       \\
             $K{=}512$&     277M    &  817M     & -       \\
            \bottomrule
        \end{tabular}
    }
    \hfill\null
    \caption{
    \textbf{Training PICs using functional sharing requires comparable resources as PCs.}
    We compare the average GPU memory (left) and time (right) required to perform an Adam optimization step with PCs (blue) and PICs (green), varying region graph type, parameterizations, and $K$.
    We pair the plots with a table reporting the number of parameters of PCs at different $K$ and PICs, with the latter being independent of $K$ and allowing up to 99\% less trainable parameters (\rgqg-\lcp-512).
    The benchmark is conducted using a batch size of 256 gray-scale images of size 28x28, i.e.\ \mnist-like.
    Extra (tabular) details in \cref{tab:bech28x28}.
    }
    \label{fig:bench28x28}
\end{figure*}

We report the time and GPU memory required to perform and Adamp optimization step for several model configurations in \cref{tab:bech28x28}, \cref{fig:bench28x28} and \cref{tab:bench64x64}.

\begin{table}[!htb]
\caption{
\textbf{Training PICs using functional sharing requires comparable resources as PCs.}
We report the time (in milliseconds, top) and GPU memory (in GiB, bottom) required to perform an Adam optimization step on 256 \mnist-like images by varying: architecture ($\{\text{QT-CP, QG-CP, QG-TK}\}$), size $K$ ($\{2^i\}_{i=4}^{9}$), model ($\{\text{PC}, \text{PIC}\}$), and sharing technique ($\{\text{C}, \text{F}, \text{N}\}$).
For each model we attach a pair $(\cdot, \cdot)$ where the first (resp.\ second) argument specifies the sharing technique for the input (resp.\ inner) layer(s).
}
\label{tab:bech28x28}
\vspace{3mm}

\centering
\begin{tabular}{cr|cc|ccc}
    RG-layer & $K$  &
    \begin{tabular}{@{}c@{}}PC   \\ (F, N) \end{tabular} &
    \begin{tabular}{@{}c@{}}PC   \\ (N, N) \end{tabular} &
    \begin{tabular}{@{}c@{}}PIC \\ (F, C) \end{tabular} &
    \begin{tabular}{@{}c@{}}PIC \\ (C, C) \end{tabular} &
    \begin{tabular}{@{}c@{}}PIC \\ (F, N) \end{tabular} \\

    \hline
    \multirow{6}{*}{\rgqt-\lcp}

    & $16$ &\cellcolor[rgb]{0.41, 0.68, 0.42} 16 & \cellcolor[rgb]{0.41, 0.68, 0.42} 16 & \cellcolor[rgb]{0.41, 0.68, 0.42} 18 & \cellcolor[rgb]{0.41, 0.68, 0.42} 18 & \cellcolor[rgb]{0.62, 0.78, 0.46} 67 \\
    & $32$ &\cellcolor[rgb]{0.41, 0.68, 0.42} 17 & \cellcolor[rgb]{0.41, 0.68, 0.42} 17 & \cellcolor[rgb]{0.41, 0.68, 0.42} 19 & \cellcolor[rgb]{0.41, 0.68, 0.42} 19 & \cellcolor[rgb]{0.62, 0.78, 0.46} 63 \\
    & $64$ &\cellcolor[rgb]{0.41, 0.68, 0.42} 19 & \cellcolor[rgb]{0.41, 0.68, 0.42} 19 & \cellcolor[rgb]{0.41, 0.68, 0.42} 21 & \cellcolor[rgb]{0.41, 0.68, 0.42} 21 & \cellcolor[rgb]{0.71, 0.83, 0.51} 89 \\
    & $128$ &\cellcolor[rgb]{0.41, 0.68, 0.42} 24 & \cellcolor[rgb]{0.41, 0.68, 0.42} 24 & \cellcolor[rgb]{0.41, 0.68, 0.42} 25 & \cellcolor[rgb]{0.41, 0.68, 0.42} 26 & OOM\\
    & $256$ &\cellcolor[rgb]{0.51, 0.73, 0.45} 38 & \cellcolor[rgb]{0.51, 0.73, 0.45} 40 & \cellcolor[rgb]{0.51, 0.73, 0.45} 43 & \cellcolor[rgb]{0.51, 0.73, 0.45} 47 & OOM\\
    & $512$ &\cellcolor[rgb]{0.71, 0.83, 0.51} 82 & \cellcolor[rgb]{0.71, 0.83, 0.51} 90 & \cellcolor[rgb]{0.79, 0.87, 0.57} 116 & \cellcolor[rgb]{0.86, 0.91, 0.65} 128 & OOM\\

    \hline
    \multirow{6}{*}{\rgqg-\lcp}

    & $16$ &\cellcolor[rgb]{0.51, 0.73, 0.45} 48 & \cellcolor[rgb]{0.51, 0.73, 0.45} 48 & \cellcolor[rgb]{0.51, 0.73, 0.45} 43 & \cellcolor[rgb]{0.51, 0.73, 0.45} 43 & \cellcolor[rgb]{0.95, 0.88, 0.65} 193 \\
    & $32$ &\cellcolor[rgb]{0.51, 0.73, 0.45} 52 & \cellcolor[rgb]{0.51, 0.73, 0.45} 52 & \cellcolor[rgb]{0.51, 0.73, 0.45} 45 & \cellcolor[rgb]{0.51, 0.73, 0.45} 46 & \cellcolor[rgb]{0.95, 0.88, 0.65} 189 \\
    & $64$ &\cellcolor[rgb]{0.51, 0.73, 0.45} 58 & \cellcolor[rgb]{0.62, 0.78, 0.46} 59 & \cellcolor[rgb]{0.51, 0.73, 0.45} 52 & \cellcolor[rgb]{0.51, 0.73, 0.45} 52 & OOM\\
    & $128$ &\cellcolor[rgb]{0.62, 0.78, 0.46} 74 & \cellcolor[rgb]{0.62, 0.78, 0.46} 74 & \cellcolor[rgb]{0.62, 0.78, 0.46} 66 & \cellcolor[rgb]{0.62, 0.78, 0.46} 67 & OOM\\
    & $256$ &\cellcolor[rgb]{0.79, 0.87, 0.57} 119 & \cellcolor[rgb]{0.79, 0.87, 0.57} 121 & \cellcolor[rgb]{0.79, 0.87, 0.57} 122 & \cellcolor[rgb]{0.86, 0.91, 0.65} 126 & OOM\\
    & $512$ &\cellcolor[rgb]{0.91, 0.65, 0.44} 257 & \cellcolor[rgb]{0.91, 0.65, 0.44} 264 & \cellcolor[rgb]{0.80, 0.35, 0.28} 326 & \cellcolor[rgb]{0.80, 0.35, 0.28} 337 & OOM\\

    \hline
    \multirow{3}{*}{\rgqg-\ltucker}

    & $16$ &\cellcolor[rgb]{0.62, 0.78, 0.46} 71 & \cellcolor[rgb]{0.62, 0.78, 0.46} 72 & \cellcolor[rgb]{0.51, 0.73, 0.45} 56 & \cellcolor[rgb]{0.51, 0.73, 0.45} 57 & \cellcolor[rgb]{0.93, 0.95, 0.73} 156 \\
    & $32$ &\cellcolor[rgb]{0.79, 0.87, 0.57} 102 & \cellcolor[rgb]{0.79, 0.87, 0.57} 102 & \cellcolor[rgb]{0.71, 0.83, 0.51} 89 & \cellcolor[rgb]{0.71, 0.83, 0.51} 89 & OOM\\
    & $64$ &\cellcolor[rgb]{0.93, 0.82, 0.57} 221 & \cellcolor[rgb]{0.93, 0.82, 0.57} 221 & \cellcolor[rgb]{0.89, 0.55, 0.39} 288 & \cellcolor[rgb]{0.89, 0.55, 0.39} 289 & OOM\\

\end{tabular}

\vspace{10mm}

\begin{tabular}{cr|cc|ccc}


    RG-layer & $K$  &
    \begin{tabular}{@{}c@{}}PC   \\ (F, N) \end{tabular} &
    \begin{tabular}{@{}c@{}}PC   \\ (N, N) \end{tabular} &
    \begin{tabular}{@{}c@{}}PIC \\ (F, C) \end{tabular} &
    \begin{tabular}{@{}c@{}}PIC \\ (C, C) \end{tabular} &
    \begin{tabular}{@{}c@{}}PIC \\ (F, N) \end{tabular} \\

    \hline
    \multirow{6}{*}{\rgqt-\lcp}

    & $16$ &\cellcolor[rgb]{0.41, 0.68, 0.42} 0.08 & \cellcolor[rgb]{0.41, 0.68, 0.42} 0.14 & \cellcolor[rgb]{0.41, 0.68, 0.42} 0.15 & \cellcolor[rgb]{0.41, 0.68, 0.42} 1.00 & \cellcolor[rgb]{0.51, 0.73, 0.45} 2.68 \\
    & $32$ &\cellcolor[rgb]{0.41, 0.68, 0.42} 0.16 & \cellcolor[rgb]{0.41, 0.68, 0.42} 0.28 & \cellcolor[rgb]{0.41, 0.68, 0.42} 0.19 & \cellcolor[rgb]{0.41, 0.68, 0.42} 1.05 & \cellcolor[rgb]{0.79, 0.87, 0.57} 6.97 \\
    & $64$ &\cellcolor[rgb]{0.41, 0.68, 0.42} 0.36 & \cellcolor[rgb]{0.41, 0.68, 0.42} 0.60 & \cellcolor[rgb]{0.41, 0.68, 0.42} 0.40 & \cellcolor[rgb]{0.41, 0.68, 0.42} 1.27 & \cellcolor[rgb]{0.86, 0.45, 0.34} 20.11 \\
    & $128$ &\cellcolor[rgb]{0.41, 0.68, 0.42} 0.83 & \cellcolor[rgb]{0.41, 0.68, 0.42} 1.31 & \cellcolor[rgb]{0.41, 0.68, 0.42} 0.97 & \cellcolor[rgb]{0.51, 0.73, 0.45} 1.93 & OOM\\
    & $256$ &\cellcolor[rgb]{0.51, 0.73, 0.45} 2.17 & \cellcolor[rgb]{0.51, 0.73, 0.45} 3.12 & \cellcolor[rgb]{0.51, 0.73, 0.45} 2.68 & \cellcolor[rgb]{0.62, 0.78, 0.46} 3.83 & OOM\\
    & $512$ &\cellcolor[rgb]{0.79, 0.87, 0.57} 6.86 & \cellcolor[rgb]{0.86, 0.91, 0.65} 8.39 & \cellcolor[rgb]{0.86, 0.91, 0.65} 8.75 & \cellcolor[rgb]{0.93, 0.95, 0.73} 10.09 & OOM\\

    \hline
    \multirow{6}{*}{\rgqg-\lcp}

    & $16$ &\cellcolor[rgb]{0.41, 0.68, 0.42} 0.18 & \cellcolor[rgb]{0.41, 0.68, 0.42} 0.24 & \cellcolor[rgb]{0.41, 0.68, 0.42} 0.22 & \cellcolor[rgb]{0.41, 0.68, 0.42} 1.03 & \cellcolor[rgb]{0.86, 0.91, 0.65} 7.85 \\
    & $32$ &\cellcolor[rgb]{0.41, 0.68, 0.42} 0.38 & \cellcolor[rgb]{0.41, 0.68, 0.42} 0.49 & \cellcolor[rgb]{0.41, 0.68, 0.42} 0.42 & \cellcolor[rgb]{0.41, 0.68, 0.42} 1.24 & \cellcolor[rgb]{0.93, 0.82, 0.57} 13.95 \\
    & $64$ &\cellcolor[rgb]{0.41, 0.68, 0.42} 0.83 & \cellcolor[rgb]{0.41, 0.68, 0.42} 1.07 & \cellcolor[rgb]{0.41, 0.68, 0.42} 0.92 & \cellcolor[rgb]{0.51, 0.73, 0.45} 1.79 & OOM\\
    & $128$ &\cellcolor[rgb]{0.51, 0.73, 0.45} 2.03 & \cellcolor[rgb]{0.51, 0.73, 0.45} 2.50 & \cellcolor[rgb]{0.51, 0.73, 0.45} 2.29 & \cellcolor[rgb]{0.62, 0.78, 0.46} 3.25 & OOM\\
    & $256$ &\cellcolor[rgb]{0.71, 0.83, 0.51} 5.54 & \cellcolor[rgb]{0.79, 0.87, 0.57} 6.50 & \cellcolor[rgb]{0.79, 0.87, 0.57} 6.66 & \cellcolor[rgb]{0.79, 0.87, 0.57} 7.66 & OOM\\
    & $512$ &\cellcolor[rgb]{0.91, 0.65, 0.44} 17.14 & \cellcolor[rgb]{0.89, 0.55, 0.39} 19.05 & \cellcolor[rgb]{0.80, 0.35, 0.28} 21.70 & \cellcolor[rgb]{0.80, 0.35, 0.28} 23.04 & OOM\\

    \hline
    \multirow{3}{*}{\rgqg-\ltucker}

    & $16$ &\cellcolor[rgb]{0.41, 0.68, 0.42} 0.68 & \cellcolor[rgb]{0.41, 0.68, 0.42} 0.73 & \cellcolor[rgb]{0.41, 0.68, 0.42} 0.80 & \cellcolor[rgb]{0.41, 0.68, 0.42} 1.55 & \cellcolor[rgb]{0.89, 0.55, 0.39} 18.60 \\
    & $32$ &\cellcolor[rgb]{0.51, 0.73, 0.45} 2.94 & \cellcolor[rgb]{0.51, 0.73, 0.45} 3.04 & \cellcolor[rgb]{0.62, 0.78, 0.46} 3.75 & \cellcolor[rgb]{0.62, 0.78, 0.46} 4.37 & OOM\\
    & $64$ &\cellcolor[rgb]{0.93, 0.82, 0.57} 14.72 & \cellcolor[rgb]{0.93, 0.82, 0.57} 14.92 & \cellcolor[rgb]{0.86, 0.45, 0.34} 21.04 & \cellcolor[rgb]{0.80, 0.35, 0.28} 21.71 & OOM\\

\end{tabular}

\end{table}

\begin{table}[!htb]
\caption{
\textbf{PICs using functional sharing require comparable resources as PCs to be trained, and scale much more gracefully than PICs without sharing.}
We report the time (in milliseconds, top) and GPU memory (in GiB, bottom) required to perform an Adam optimization step on a batch of 128 RGB images of size 64x64 by varying: architecture ($\{\text{QT-CP, QG-CP, QG-TK}\}$), size $K$ ($\{2^i\}_{i=4}^{8}$), model ($\{\text{PC}, \text{PIC}\}$), sharing technique ($\{\text{C}, \text{F}, \text{N}\}$), and also the MLP size $M$ ($\{2^i\}_{i=5}^{8}$) of the PIC models.
For each model we attach a pair $(\cdot, \cdot)$ where the first (resp. second) argument specifies the sharing technique for the input (resp. inner) layer(s).
}
\label{tab:bench64x64}
\vspace{3mm}

\setlength{\tabcolsep}{3.4pt}
\centering
\begin{tabular}{cr|c|cccc|cccc}

    \multirow{2}{*}{} &
    \multirow{2}{*}[0pt]{\vspace{-1.5ex}$K$} &
    \multirow{2}{*}[0pt]{\vspace{-1.5ex}PC (F, N)} &
    \multicolumn{4}{c}{PIC (F, C)} &
    \multicolumn{4}{c}{PIC (F, N)} \\
    & & & $M\,{=}\,256$ & $128$ & $64$ & $32$ & $M\,{=}\,256$ & $128$ & $64$ & $32$ \\

    \hline
    \multirow{5}{*}{\rgqt-\lcp}

    & $16$ &\cellcolor[rgb]{0.41, 0.68, 0.42} 37 & \cellcolor[rgb]{0.41, 0.68, 0.42} 40 & \cellcolor[rgb]{0.41, 0.68, 0.42} 38 & \cellcolor[rgb]{0.41, 0.68, 0.42} 40 & \cellcolor[rgb]{0.41, 0.68, 0.42} 39 & \cellcolor[rgb]{0.93, 0.95, 0.73} 330 & \cellcolor[rgb]{0.71, 0.83, 0.51} 203 & \cellcolor[rgb]{0.51, 0.73, 0.45} 94 & \cellcolor[rgb]{0.41, 0.68, 0.42} 59 \\
    & $32$ &\cellcolor[rgb]{0.41, 0.68, 0.42} 51 & \cellcolor[rgb]{0.41, 0.68, 0.42} 53 & \cellcolor[rgb]{0.41, 0.68, 0.42} 49 & \cellcolor[rgb]{0.41, 0.68, 0.42} 53 & \cellcolor[rgb]{0.41, 0.68, 0.42} 51 & \cellcolor[rgb]{0.96, 0.94, 0.73} 388 & \cellcolor[rgb]{0.62, 0.78, 0.46} 161 & \cellcolor[rgb]{0.51, 0.73, 0.45} 124 & \cellcolor[rgb]{0.41, 0.68, 0.42} 72 \\
    & $64$ &\cellcolor[rgb]{0.41, 0.68, 0.42} 74 & \cellcolor[rgb]{0.41, 0.68, 0.42} 78 & \cellcolor[rgb]{0.41, 0.68, 0.42} 77 & \cellcolor[rgb]{0.41, 0.68, 0.42} 77 & \cellcolor[rgb]{0.41, 0.68, 0.42} 76 & OOM &OOM &\cellcolor[rgb]{0.79, 0.87, 0.57} 245 & \cellcolor[rgb]{0.51, 0.73, 0.45} 120 \\
    & $128$ &\cellcolor[rgb]{0.51, 0.73, 0.45} 126 & \cellcolor[rgb]{0.51, 0.73, 0.45} 126 & \cellcolor[rgb]{0.51, 0.73, 0.45} 124 & \cellcolor[rgb]{0.51, 0.73, 0.45} 126 & \cellcolor[rgb]{0.51, 0.73, 0.45} 126 & OOM &OOM &OOM &OOM\\
    & $256$ &\cellcolor[rgb]{0.79, 0.87, 0.57} 246 & \cellcolor[rgb]{0.79, 0.87, 0.57} 257 & \cellcolor[rgb]{0.79, 0.87, 0.57} 253 & \cellcolor[rgb]{0.79, 0.87, 0.57} 249 & \cellcolor[rgb]{0.79, 0.87, 0.57} 251 & OOM &OOM &OOM &OOM\\

    \hline
    \multirow{5}{*}{\rgqg-\lcp}

    & $16$ &\cellcolor[rgb]{0.41, 0.68, 0.42} 77 & \cellcolor[rgb]{0.41, 0.68, 0.42} 73 & \cellcolor[rgb]{0.41, 0.68, 0.42} 71 & \cellcolor[rgb]{0.41, 0.68, 0.42} 71 & \cellcolor[rgb]{0.41, 0.68, 0.42} 70 & OOM &\cellcolor[rgb]{0.92, 0.74, 0.51} 550 & \cellcolor[rgb]{0.79, 0.87, 0.57} 238 & \cellcolor[rgb]{0.51, 0.73, 0.45} 129 \\
    & $32$ &\cellcolor[rgb]{0.51, 0.73, 0.45} 98 & \cellcolor[rgb]{0.51, 0.73, 0.45} 94 & \cellcolor[rgb]{0.51, 0.73, 0.45} 92 & \cellcolor[rgb]{0.51, 0.73, 0.45} 88 & \cellcolor[rgb]{0.51, 0.73, 0.45} 89 & OOM &\cellcolor[rgb]{0.80, 0.35, 0.28} 762 & \cellcolor[rgb]{0.86, 0.91, 0.65} 312 & \cellcolor[rgb]{0.62, 0.78, 0.46} 150 \\
    & $64$ &\cellcolor[rgb]{0.62, 0.78, 0.46} 135 & \cellcolor[rgb]{0.51, 0.73, 0.45} 132 & \cellcolor[rgb]{0.51, 0.73, 0.45} 130 & \cellcolor[rgb]{0.51, 0.73, 0.45} 128 & \cellcolor[rgb]{0.51, 0.73, 0.45} 123 & OOM &OOM &OOM &\cellcolor[rgb]{0.79, 0.87, 0.57} 264 \\
    & $128$ &\cellcolor[rgb]{0.71, 0.83, 0.51} 215 & \cellcolor[rgb]{0.71, 0.83, 0.51} 213 & \cellcolor[rgb]{0.71, 0.83, 0.51} 209 & \cellcolor[rgb]{0.71, 0.83, 0.51} 208 & \cellcolor[rgb]{0.71, 0.83, 0.51} 201 & OOM &OOM &OOM &OOM\\
    & $256$ &\cellcolor[rgb]{0.95, 0.88, 0.65} 428 & \cellcolor[rgb]{0.95, 0.88, 0.65} 449 & \cellcolor[rgb]{0.95, 0.88, 0.65} 440 & \cellcolor[rgb]{0.95, 0.88, 0.65} 433 & \cellcolor[rgb]{0.95, 0.88, 0.65} 425 & OOM &OOM &OOM &OOM\\

    \hline
    \multirow{2}{*}{\rgqg-\ltucker}

    & $16$ &\cellcolor[rgb]{0.51, 0.73, 0.45} 117 & \cellcolor[rgb]{0.51, 0.73, 0.45} 99 & \cellcolor[rgb]{0.51, 0.73, 0.45} 98 & \cellcolor[rgb]{0.51, 0.73, 0.45} 98 & \cellcolor[rgb]{0.51, 0.73, 0.45} 95 & OOM &OOM &\cellcolor[rgb]{0.95, 0.88, 0.65} 429 & \cellcolor[rgb]{0.62, 0.78, 0.46} 164 \\
    & $32$ &\cellcolor[rgb]{0.71, 0.83, 0.51} 211 & \cellcolor[rgb]{0.71, 0.83, 0.51} 204 & \cellcolor[rgb]{0.71, 0.83, 0.51} 202 & \cellcolor[rgb]{0.71, 0.83, 0.51} 199 & \cellcolor[rgb]{0.71, 0.83, 0.51} 183 & OOM &OOM &OOM &OOM\\

\end{tabular}

\vspace{10mm}

\centering
\begin{tabular}{cr|c|cccc|cccc}

    \multirow{2}{*}{} &
    \multirow{2}{*}[0pt]{\vspace{-1.5ex}$K$} &
    \multirow{2}{*}[0pt]{\vspace{-1.5ex}PC (F, N)} &
    \multicolumn{4}{c}{PIC (F, C)} &
    \multicolumn{4}{c}{PIC (F, N)} \\
    & & & $M\,{=}\,256$ & $128$ & $64$ & $32$ & $M\,{=}\,256$ & $128$ & $64$ & $32$ \\

    \hline
    \multirow{5}{*}{\rgqt-\lcp}

    & $16$ &\cellcolor[rgb]{0.41, 0.68, 0.42} 0.67 & \cellcolor[rgb]{0.41, 0.68, 0.42} 0.70 & \cellcolor[rgb]{0.41, 0.68, 0.42} 0.68 & \cellcolor[rgb]{0.41, 0.68, 0.42} 0.67 & \cellcolor[rgb]{0.41, 0.68, 0.42} 0.67 & \cellcolor[rgb]{0.93, 0.95, 0.73} 13.81 & \cellcolor[rgb]{0.62, 0.78, 0.46} 5.03 & \cellcolor[rgb]{0.41, 0.68, 0.42} 2.14 & \cellcolor[rgb]{0.41, 0.68, 0.42} 1.13 \\
    & $32$ &\cellcolor[rgb]{0.41, 0.68, 0.42} 1.23 & \cellcolor[rgb]{0.41, 0.68, 0.42} 1.27 & \cellcolor[rgb]{0.41, 0.68, 0.42} 1.25 & \cellcolor[rgb]{0.41, 0.68, 0.42} 1.24 & \cellcolor[rgb]{0.41, 0.68, 0.42} 1.23 & \cellcolor[rgb]{0.80, 0.35, 0.28} 31.45 & \cellcolor[rgb]{0.86, 0.91, 0.65} 12.54 & \cellcolor[rgb]{0.62, 0.78, 0.46} 6.15 & \cellcolor[rgb]{0.51, 0.73, 0.45} 2.94 \\
    & $64$ &\cellcolor[rgb]{0.41, 0.68, 0.42} 2.48 & \cellcolor[rgb]{0.41, 0.68, 0.42} 2.57 & \cellcolor[rgb]{0.41, 0.68, 0.42} 2.52 & \cellcolor[rgb]{0.41, 0.68, 0.42} 2.50 & \cellcolor[rgb]{0.41, 0.68, 0.42} 2.49 & OOM &OOM &\cellcolor[rgb]{0.92, 0.74, 0.51} 22.59 & \cellcolor[rgb]{0.86, 0.91, 0.65} 10.97 \\
    & $128$ &\cellcolor[rgb]{0.62, 0.78, 0.46} 5.48 & \cellcolor[rgb]{0.62, 0.78, 0.46} 5.74 & \cellcolor[rgb]{0.62, 0.78, 0.46} 5.61 & \cellcolor[rgb]{0.62, 0.78, 0.46} 5.54 & \cellcolor[rgb]{0.62, 0.78, 0.46} 5.51 & OOM &OOM &OOM &OOM\\
    & $256$ &\cellcolor[rgb]{0.93, 0.95, 0.73} 13.48 & \cellcolor[rgb]{0.93, 0.95, 0.73} 14.45 & \cellcolor[rgb]{0.93, 0.95, 0.73} 13.96 & \cellcolor[rgb]{0.93, 0.95, 0.73} 13.72 & \cellcolor[rgb]{0.93, 0.95, 0.73} 13.60 & OOM &OOM &OOM &OOM\\

    \hline
    \multirow{5}{*}{\rgqg-\lcp}

    & $16$ &\cellcolor[rgb]{0.41, 0.68, 0.42} 0.71 & \cellcolor[rgb]{0.41, 0.68, 0.42} 0.78 & \cellcolor[rgb]{0.41, 0.68, 0.42} 0.74 & \cellcolor[rgb]{0.41, 0.68, 0.42} 0.72 & \cellcolor[rgb]{0.41, 0.68, 0.42} 0.72 & OOM &\cellcolor[rgb]{0.86, 0.91, 0.65} 12.68 & \cellcolor[rgb]{0.62, 0.78, 0.46} 4.88 & \cellcolor[rgb]{0.41, 0.68, 0.42} 2.10 \\
    & $32$ &\cellcolor[rgb]{0.41, 0.68, 0.42} 1.40 & \cellcolor[rgb]{0.41, 0.68, 0.42} 1.50 & \cellcolor[rgb]{0.41, 0.68, 0.42} 1.44 & \cellcolor[rgb]{0.41, 0.68, 0.42} 1.42 & \cellcolor[rgb]{0.41, 0.68, 0.42} 1.41 & OOM &\cellcolor[rgb]{0.86, 0.45, 0.34} 27.72 & \cellcolor[rgb]{0.86, 0.91, 0.65} 12.93 & \cellcolor[rgb]{0.62, 0.78, 0.46} 6.00 \\
    & $64$ &\cellcolor[rgb]{0.51, 0.73, 0.45} 3.15 & \cellcolor[rgb]{0.51, 0.73, 0.45} 3.35 & \cellcolor[rgb]{0.51, 0.73, 0.45} 3.24 & \cellcolor[rgb]{0.51, 0.73, 0.45} 3.20 & \cellcolor[rgb]{0.51, 0.73, 0.45} 3.17 & OOM &OOM &OOM &\cellcolor[rgb]{0.92, 0.74, 0.51} 22.13 \\
    & $128$ &\cellcolor[rgb]{0.71, 0.83, 0.51} 8.15 & \cellcolor[rgb]{0.71, 0.83, 0.51} 8.74 & \cellcolor[rgb]{0.71, 0.83, 0.51} 8.44 & \cellcolor[rgb]{0.71, 0.83, 0.51} 8.30 & \cellcolor[rgb]{0.71, 0.83, 0.51} 8.22 & OOM &OOM &OOM &OOM\\
    & $256$ &\cellcolor[rgb]{0.91, 0.65, 0.44} 24.14 & \cellcolor[rgb]{0.89, 0.55, 0.39} 26.29 & \cellcolor[rgb]{0.91, 0.65, 0.44} 25.22 & \cellcolor[rgb]{0.91, 0.65, 0.44} 24.69 & \cellcolor[rgb]{0.91, 0.65, 0.44} 24.42 & OOM &OOM &OOM &OOM\\

    \hline
    \multirow{2}{*}{\rgqg-\ltucker}

    & $16$ &\cellcolor[rgb]{0.41, 0.68, 0.42} 2.24 & \cellcolor[rgb]{0.41, 0.68, 0.42} 2.35 & \cellcolor[rgb]{0.41, 0.68, 0.42} 2.26 & \cellcolor[rgb]{0.41, 0.68, 0.42} 2.22 & \cellcolor[rgb]{0.41, 0.68, 0.42} 2.20 & OOM &OOM &\cellcolor[rgb]{0.92, 0.74, 0.51} 22.46 & \cellcolor[rgb]{0.86, 0.91, 0.65} 11.10 \\
    & $32$ &\cellcolor[rgb]{0.79, 0.87, 0.57} 10.77 & \cellcolor[rgb]{0.86, 0.91, 0.65} 11.35 & \cellcolor[rgb]{0.79, 0.87, 0.57} 10.81 & \cellcolor[rgb]{0.79, 0.87, 0.57} 10.54 & \cellcolor[rgb]{0.79, 0.87, 0.57} 10.40 & OOM &OOM &OOM &OOM\\

\end{tabular}

\end{table}

\clearpage
\subsection{Additional distribution estimation results}
\label{app:distribution-estimation}

In this section, we report tabular results for all our experiments.

Note that, every input layer of a standard PC is parameterized by a matrix $K \, {\times} \, P$, where $P$ is the number of categories, which is $256$ for grey-scale image datasets and $768 \, {=} 256 \cdot 3$ for RGB images.
We found that sharing a single input layer among all pixels results in (slightly) worse performance for grey-scale images (as detailed in \cref{tab:mnist-pc-sharing}).
In contrast, we found that such sharing considerably improves performance for RGB image datasets.
Besides improving performance for RGB image datasets, such sharing considerably lower the number of trainable parameters from $D \, {\times} \, K \, {\times} \, P$ to only $K \, {\times} \, P$ where $D$ is the number of pixels.
For instance, parameterizing all input layers of a tensorized architecture with $K \, {=} \, 256$ built for 64x64 images would require $805,306,368 = 256 \, {\cdot} \, 64 \, {\cdot} \, 64 \, {\cdot} \, 768$ parameters, while only $3,145,728$ if we apply the sharing.
Therefore, without applying such sharing, we cannot even scale to big tensorized architectures (e.g.\ \rgqg-\lcp-512) on our GPUs.

All QPCs are materialized from PICs applying F-sharing over the group of input units, and C-sharing over the groups of integral units.

We extensively compare QPCs and PCs as density estimators on several image datasets and report test-set bits-per-dimension (bpd) in \cref{tab:mnist-pc-sharing}, \cref{tab:mnist-fmnist-ablation}, and \cref{tab:28x28_32x32_64x64}.

\begin{table}[!htb]
\centering
\caption{
\textbf{PCs with a shared input layer deliver comparable performance as PCs who do not on the \mnist-family datasets.}
We compare the bits-per-dimension of PCs with (w/) and without (w/o) a shared input layer considering three different tensorized architectures: \rgqt-\lcp-512, \rgqg-\lcp-512 and \rgqg-\ltucker-64.
}
\label{tab:mnist-pc-sharing}
\vspace{3mm}
    \small
    \setlength{\tabcolsep}{3pt}
    \centering

    \begin{tabular}{r|cc|cc|cc}

    \multirow{2}{*}{} & \multicolumn{2}{c|}{\rgqt-\lcp-512} & \multicolumn{2}{c|}{\rgqg-\lcp-512} & \multicolumn{2}{c}{\rgqg-\ltucker-64} \\
    & w/o & w/ & w/o & w/ & w/o & w/ \\
    \toprule

    \textsc{mnist}      & $\textbf{1.175} \pm 0.001$    & $1.213 \pm 0.002$             & $\textbf{1.177} \pm 0.006$     & $1.241 \pm 0.005$     & $\textbf{1.257} \pm 0.005$   & $1.300 \pm 0.004$ \\
    \textsc{f-mnist}    & $\textbf{3.381} \pm 0.001$    & $\textbf{3.381} \pm 0.002$    & $\textbf{3.317} \pm 0.005$     & $3.375 \pm 0.005$     & $\textbf{3.499} \pm 0.006$   & $3.560 \pm 0.007$ \\
    \textsc{emn-mn}     & $\textbf{1.706} \pm 0.007$    & $1.761 \pm 0.005$             & $\textbf{1.643} \pm 0.007$     & $1.711 \pm 0.006$     & $\textbf{1.756} \pm 0.002$   & $1.772 \pm 0.004$ \\
    \textsc{emn-le}     & $\textbf{1.698} \pm 0.006$    & $1.735 \pm 0.007$             & $\textbf{1.626} \pm 0.004$     & $1.656 \pm 0.004$     & $\textbf{1.725} \pm 0.003$   & $1.728 \pm 0.003$ \\
    \textsc{emn-ba}     & $1.731 \pm 0.007$             & $\textbf{1.772} \pm 0.007$    & $\textbf{1.665} \pm 0.004$     & $1.696 \pm 0.003$     & $1.751 \pm 0.002$            & $\textbf{1.749} \pm 0.005$ \\
    \textsc{emn-by}     & $\textbf{1.542} \pm 0.008$    & $1.548 \pm 0.007$             & $\textbf{1.474} \pm 0.009$     & $1.481 \pm 0.007$     & $\textbf{1.665} \pm 0.007$   & $1.679 \pm 0.006$ \\
    \end{tabular}
\end{table}

\begin{table}[!htb]
\centering
\caption{
\textbf{QPCs consistently improve over PCs on \mnist and \fmnist.}
Test-set bits-per-dimension (bpd) on \mnist (top) and \fmnist (bottom) averaged over 5 runs.
All QPCs are materialized from PICs applying F-sharing over the group of input units, and C-sharing over the integral units groups, i.e.\ QPCs are materialized from PICs (F, C).
PCs do not apply any form of parameter sharing, as these deliver the best performance for these datasets, as detailed in \cref{tab:mnist-pc-sharing}.
}
\vspace{3mm}
\label{tab:mnist-fmnist-ablation}
    \small
    \setlength{\tabcolsep}{4pt}
    \centering
    \begin{tabular}{r|cc|cc|cc}

    \multirow{2}{*}{$K$} & \multicolumn{2}{c|}{\rgqt-\lcp} & \multicolumn{2}{c|}{\rgqg-\lcp} & \multicolumn{2}{c}{\rgqg-\ltucker} \\
    & QPC & PC & QPC & PC & QPC & PC \\

    \toprule

    $16$        & $\textbf{1.275} \pm 0.009$     & $\textbf{1.283} \pm 0.004$     & $\textbf{1.237} \pm 0.009$    & $1.248 \pm 0.003$      & $\textbf{1.215} \pm 0.010$ & $1.233 \pm 0.004$   \\
    $32$        & $\textbf{1.220} \pm 0.003$     & $\textbf{1.242} \pm 0.004$     & $\textbf{1.189} \pm 0.008$    & $1.212 \pm 0.003$      & $\textbf{1.168} \pm 0.002$ & $1.222 \pm 0.004$   \\
    $64$        & $\textbf{1.195} \pm 0.002$     & $\textbf{1.217} \pm 0.002$     & $\textbf{1.162} \pm 0.002$    & $1.185 \pm 0.002$      & $\textbf{1.144} \pm 0.006$ & $1.257 \pm 0.005$   \\
    $128$       & $\textbf{1.161} \pm 0.003$     & $\textbf{1.196} \pm 0.004$     & $\textbf{1.144} \pm 0.006$    & $1.171 \pm 0.002$      & \multicolumn{2}{c}{OOM} \\
    $256$       & $\textbf{1.135} \pm 0.006$     & $\textbf{1.184} \pm 0.002$     & $\textbf{1.120} \pm 0.005$    & $1.173 \pm 0.009$      & \multicolumn{2}{c}{OOM} \\
    $512$       & $\textbf{1.126} \pm 0.004$     & $\textbf{1.175} \pm 0.001$     & $\textbf{1.115} \pm 0.005$    & $1.177 \pm 0.006$      & \multicolumn{2}{c}{OOM} \\

    \end{tabular}

    \vspace{5mm}

    \small
    \setlength{\tabcolsep}{4pt}
    \centering
    \begin{tabular}{r|cc|cc|cc}

    \toprule

    $16$        & $\textbf{3.547} \pm 0.003$     & $3.589 \pm 0.005$     & $\textbf{3.427} \pm 0.006$    & $3.464 \pm 0.005$      & $\textbf{3.424} \pm 0.009$  & $3.446 \pm 0.008$ \\
    $32$        & $\textbf{3.429} \pm 0.001$     & $3.497 \pm 0.003$     & $\textbf{3.319} \pm 0.001$    & $3.385 \pm 0.004$      & $\textbf{3.323} \pm 0.003$  & $3.417 \pm 0.005$ \\
    $64$        & $\textbf{3.349} \pm 0.005$     & $3.442 \pm 0.003$     & $\textbf{3.258} \pm 0.004$    & $3.339 \pm 0.004$      & $\textbf{3.251} \pm 0.003$  & $3.499 \pm 0.006$ \\
    $128$       & $\textbf{3.289} \pm 0.001$     & $3.408 \pm 0.003$     & $\textbf{3.212} \pm 0.003$    & $3.319 \pm 0.004$      & \multicolumn{2}{c}{OOM} \\
    $256$       & $\textbf{3.242} \pm 0.001$     & $3.392 \pm 0.002$     & $\textbf{3.174} \pm 0.002$    & $3.317 \pm 0.002$      & \multicolumn{2}{c}{OOM} \\
    $512$       & $\textbf{3.209} \pm 0.003$     & $3.381 \pm 0.001$     & $\textbf{3.154} \pm 0.004$    & $3.317 \pm 0.005$      & \multicolumn{2}{c}{OOM} \\

    \end{tabular}

\end{table}

\clearpage

\begin{table}[!htb]
\centering
\caption[Test-set bits-per-dimensions of the largest (Q)PCs models on several image datasets.]{
\textbf{QPCs generally improve over PCs on distribution estimation.}
We report the average test-set bits-per-dimensions of \rgqt-\lcp-512, \rgqg-\lcp-512 and \rgqg-\ltucker-64 for datasets up to image size 32x32, and of \rgqt-\lcp-256, \rgqg-\lcp-256 and \rgqg-\ltucker-32 for datasets of image size 64x64.
All architectures are trained both as QPCs and PCs.
QPCs are materialized from PICs applying F-sharing over the group of input units, and C-sharing over the groups of integral units.
PCs do not apply any form of parameter sharing for \mnist-family datasets, as these delivered the best performance for such datasets \cref{tab:mnist-pc-sharing}, while they apply F-sharing at the input layer for RGB datasets.
We mark with * (resp. $\dagger$) datasets preprocessed using YCoCg-R (resp. YCoCg).
All results are averaged over 5 different runs.
}
\vspace{3mm}
\label{tab:28x28_32x32_64x64}
    \small
    \setlength{\tabcolsep}{3pt}
    \centering
    \begin{tabular}{r|cc|cc|cc}

    & QPC & PC & QPC & PC & QPC & PC \\
    \multirow{2}{*}{} & \multicolumn{2}{c|}{\rgqt-\lcp-512} & \multicolumn{2}{c|}{\rgqg-\lcp-512} & \multicolumn{2}{c}{\rgqg-\ltucker-64} \\

    \toprule

    \textsc{mnist}      & $\textbf{1.126} \pm 0.004$    & $1.175 \pm 0.001$     & $\textbf{1.115} \pm 0.005$    & $1.177 \pm 0.006$     & $\textbf{1.144} \pm 0.006$    & $1.257 \pm 0.005$ \\
    \textsc{f-mnist}    & $\textbf{3.209} \pm 0.003$    & $3.381 \pm 0.001$     & $\textbf{3.154} \pm 0.004$    & $3.317 \pm 0.005$     & $\textbf{3.251} \pm 0.003$    & $3.499 \pm 0.006$ \\
    \textsc{emn-mn}     & $\textbf{1.592} \pm 0.007$    & $1.706 \pm 0.007$     & $\textbf{1.556} \pm 0.006$    & $1.643 \pm 0.007$     & $\textbf{1.699} \pm 0.004$    & $1.756 \pm 0.002$ \\
    \textsc{emn-le}     & $\textbf{1.622} \pm 0.007$    & $1.698 \pm 0.006$     & $\textbf{1.545} \pm 0.007$    & $1.626 \pm 0.004$     & $\textbf{1.696} \pm 0.003$    & $1.725 \pm 0.003$ \\
    \textsc{emn-ba}     & $\textbf{1.638} \pm 0.006$    & $1.731 \pm 0.007$     & $\textbf{1.593} \pm 0.005$    & $1.665 \pm 0.004$     & $\textbf{1.715} \pm 0.005$    & $1.751 \pm 0.002$ \\
    \textsc{emn-by}     & $1.597 \pm 0.006$     & $\textbf{1.542} \pm 0.008$    & $1.537 \pm 0.004$     & $\textbf{1.474} \pm 0.009$    & $1.703 \pm 0.004$     & $\textbf{1.665} \pm 0.007$ \\

    $\text{CIFAR}^{*}$          & $\textbf{5.198} \pm 0.003$    & $5.596 \pm 0.004$     & $\textbf{5.097} \pm 0.002$    & $5.496 \pm 0.004$     & $\textbf{5.556} \pm 0.003$    & $5.647 \pm 0.004$ \\
    $\text{CIFAR}^{\dagger}$    & $\textbf{4.577} \pm 0.004$    & $4.884 \pm 0.003$     & $\textbf{4.486} \pm 0.009$    & $4.856 \pm 0.010$     & $\textbf{4.888} \pm 0.007$    & $4.983 \pm 0.004$ \\

    $\text{ImgNet32}^{*}$         & $\textbf{5.196} \pm 0.007$    & $5.286 \pm 0.001$     & $\textbf{5.085} \pm 0.001$    & $5.255 \pm 0.001$     & $\textbf{5.544} \pm 0.001$    & $5.700 \pm 0.002$ \\
    $\text{ImgNet32}^{\dagger}$   & $\textbf{4.578} \pm 0.001$    & $4.662 \pm 0.002$     & $\textbf{4.468} \pm 0.001$    & $4.632 \pm 0.003$     & $\textbf{4.893} \pm 0.004$    & $5.045 \pm 0.001$ \\

    &&&&&&\\
    \multirow{2}{*}{} & \multicolumn{2}{c|}{\rgqt-\lcp-256} & \multicolumn{2}{c|}{\rgqg-\lcp-256} & \multicolumn{2}{c}{\rgqg-\ltucker-32} \\
    \toprule

    $\text{CelebA}^*$               & $\textbf{4.810} \pm 0.004$    & $4.851 \pm 0.002$     & $\textbf{4.739} \pm 0.002$    &  $4.781 \pm 0.002$    & $\textbf{5.352} \pm 0.002$    &  $5.364 \pm 0.002$ \\
    $\text{CelebA}^{\dagger}$       & $\textbf{4.159} \pm 0.003$    & $4.215 \pm 0.003$     & $\textbf{4.114} \pm 0.003$    &  $4.155 \pm 0.006$    & $4.720 \pm 0.003$             &  $\textbf{5.718} \pm 0.001$ \\

    $\text{ImgNet64}^*$             & $\textbf{5.143} \pm 0.003$    & $5.221 \pm 0.003$     & $\textbf{5.051} \pm 0.002$    &  $5.220 \pm 0.005$    & $\textbf{5.657} \pm 0.004$    &  $5.764 \pm 0.001$ \\
    $\text{ImgNet64}^{\dagger}$     & $\textbf{4.523} \pm 0.003$    & $4.591 \pm 0.002$     & $\textbf{4.425} \pm 0.004$    &  $4.590 \pm 0.006$    & $\textbf{5.011} \pm 0.007$    &  $5.138 \pm 0.001$ \\

    \end{tabular}
\end{table}

\clearpage
\newpage
\section*{NeurIPS Paper Checklist}

\begin{enumerate}

\item {\bf Claims}
    \item[] Question: Do the main claims made in the abstract and introduction accurately reflect the paper's contributions and scope?
    \item[] Answer: \answerYes{}
    \item[] Justification: Abstract and main text accurately and precisely state the actual claims of the research presented.

\item {\bf Limitations}
    \item[] Question: Does the paper discuss the limitations of the work performed by the authors?
    \item[] Answer: \answerYes{}
    \item[] Justification: Limitations mentioned in the last section.

\item {\bf Theory Assumptions and Proofs}
    \item[] Question: For each theoretical result, does the paper provide the full set of assumptions and a complete (and correct) proof?
    \item[] Answer: \answerNA{}
    \item[] Justification: The paper does not include theoretical results.

    \item {\bf Experimental Result Reproducibility}
    \item[] Question: Does the paper fully disclose all the information needed to reproduce the main experimental results of the paper to the extent that it affects the main claims and/or conclusions of the paper (regardless of whether the code and data are provided or not)?
    \item[] Answer: \answerYes{}
    \item[] Justification: All data used is publicly available. All model and experimental code  to fully reproduce our own experimental results is shared.

\item {\bf Open access to data and code}
    \item[] Question: Does the paper provide open access to the data and code, with sufficient instructions to faithfully reproduce the main experimental results, as described in supplemental material?
    \item[] Answer: \answerYes{}
    \item[] Justification: All data used is publicly available. All model and experimental code  to fully reproduce our own experimental results is shared.

\item {\bf Experimental Setting/Details}
    \item[] Question: Does the paper specify all the training and test details (e.g., data splits, hyperparameters, how they were chosen, type of optimizer, etc.) necessary to understand the results?
    \item[] Answer: \answerYes{}
    \item[] Justification: Experimental settings overview included in paper; key details included in appendix; full details in the included code.

\item {\bf Experiment Statistical Significance}
    \item[] Question: Does the paper report error bars suitably and correctly defined or other appropriate information about the statistical significance of the experiments?
    \item[] Answer: \answerYes{}
    \item[] Justification: Statistical significance indicated for all experimental results, including the nature of the error bars.

\item {\bf Experiments Compute Resources}
    \item[] Question: For each experiment, does the paper provide sufficient information on the computer resources (type of compute workers, memory, time of execution) needed to reproduce the experiments?
    \item[] Answer: \answerYes{}
    \item[] Justification: Overview compute resources included in paper; key details discussed in appendix; full details listed in structured template-description of corresponding assets.

\item {\bf Code Of Ethics}
    \item[] Question: Does the research conducted in the paper conform, in every respect, with the NeurIPS Code of Ethics \url{https://neurips.cc/public/EthicsGuidelines}?
    \item[] Answer: \answerYes{}
    \item[] Justification: Problematic aspects of generative AI mentioned, used asset licenses mentioned, produced assets (code, models) licensed and shared, reporducibility ensured.

\item {\bf Broader Impacts}
    \item[] Question: Does the paper discuss both potential positive societal impacts and negative societal impacts of the work performed?
    \item[] Answer: \answerYes{}
    \item[] Justification: Positive: better interpretability of this class of generative AI models mentioned; Negative: problematic aspects of generative AI models in general mentioned.

\item {\bf Safeguards}
    \item[] Question: Does the paper describe safeguards that have been put in place for responsible release of data or models that have a high risk for misuse (e.g., pretrained language models, image generators, or scraped datasets)?
    \item[] Answer: \answerNA{}.
    \item[] Justification: The paper poses no such risks.

\item {\bf Licenses for existing assets}
    \item[] Question: Are the creators or original owners of assets (e.g., code, data, models), used in the paper, properly credited and are the license and terms of use explicitly mentioned and properly respected?
    \item[] Answer: \answerYes{}
    \item[] Justification: Original papers cited and/or URLs provided; license mentioned in reference.

\item {\bf New Assets}
    \item[] Question: Are new assets introduced in the paper well documented and is the documentation provided alongside the assets?
    \item[] Answer: \answerYes{}
    \item[] Justification: Code and models included (as zip file for submission, also as URL for final); Structured templates used for details.

\item {\bf Crowdsourcing and Research with Human Subjects}
    \item[] Question: For crowdsourcing experiments and research with human subjects, does the paper include the full text of instructions given to participants and screenshots, if applicable, as well as details about compensation (if any)?
    \item[] Answer: \answerNA{}
    \item[] Justification: No crowdsourcing or research with human subjects.

\item {\bf Institutional Review Board (IRB) Approvals or Equivalent for Research with Human Subjects}
    \item[] Question: Does the paper describe potential risks incurred by study participants, whether such risks were disclosed to the subjects, and whether Institutional Review Board (IRB) approvals (or an equivalent approval/review based on the requirements of your country or institution) were obtained?
    \item[] Answer: \answerNA{}
    \item[] Justification: No crowdsourcing or research with human subjects.

\end{enumerate}

\end{document}